\pdfoutput=1
\documentclass{article}

\usepackage[main,final]{neurips_2026}

\usepackage{booktabs}
\usepackage{amsfonts}
\usepackage{microtype}
\usepackage{xcolor}

\usepackage{tabularx}
\usepackage{array}
\newcolumntype{Y}{>{\raggedright\arraybackslash}X}
\usepackage{amsmath,amsthm}

\usepackage{algorithm}
\usepackage{algpseudocode}

\usepackage{graphicx}
\usepackage{tikz}
\usetikzlibrary{arrows.meta, backgrounds}

\usepackage{enumitem}

\usepackage{url}

\newtheorem{definition}{Definition}
\newtheorem{lemma}{Lemma}
\newtheorem{proposition}{Proposition}
\newtheorem{corollary}{Corollary}

\DeclareMathOperator*{\argmin}{arg\,min}
\DeclareMathOperator*{\argmax}{arg\,max}
\DeclareMathOperator{\RuntimeES}{RuntimeES}
\DeclareMathOperator{\Propose}{Propose}
\DeclareMathOperator{\Project}{Project}
\DeclareMathOperator{\OuterGate}{OuterGate}
\DeclareMathOperator{\StackUpdate}{StackUpdate}

\newcommand{\agentcbo}{\textsc{AgentCBO}}
\newcommand{\agentcboprior}{\textsc{AgentCBO-Prior}}
\newcommand{\agentcbocalib}{\textsc{AgentCBO-Calib}}
\newcommand{\agentcbofull}{\textsc{AgentCBO-Full}}
\newcommand{\prioragent}{\textsc{Language-Prior Agent}}
\newcommand{\calibagent}{\textsc{Calibration Agent}}
\newcommand{\repairagent}{\textsc{Residual Repair Agent}}
\newcommand{\commitgate}{\textsc{Commit Gate}}

\title{AgentCBO: Causal Belief Routing for Bayesian Optimization under Unknown Graphs}

\author{%
Qiyu Wei$^{1}$ \quad Xin Zhang$^{1}$ \quad
Richard Allmendinger$^{2}$ \quad Mauricio A. \'{A}lvarez$^{1}$\\
{\normalfont $^{1}$Department of Computer Science}\\
{\normalfont $^{2}$Alliance Manchester Business School}\\
{\normalfont The University of Manchester, Manchester, UK}\\
{\normalfont\small\texttt{\{qiyu.wei,xin.zhang-41\}@postgrad.manchester.ac.uk}}\\
{\normalfont\small\texttt{\{richard.allmendinger,mauricio.alvarezlopez\}@manchester.ac.uk}}
}

\begin{document}

\maketitle

\begin{abstract}
Causal Bayesian optimization (CBO) typically relies on the unrealistic assumption of a perfectly known causal graph. Recent methods relax this constraint by coupling optimization with structure learning, but they often waste sample budgets on structural uncertainties that are irrelevant to the optimizer. In this paper, we formalize \emph{routing-equivalence} to shift the objective from full causal-graph recovery to causal belief routing, where structural differences only matter when they alter optimizer-facing objects such as admissible exploration sets and \emph{do}-priors. We establish conditional fixed-update decision preservation and quantify CEI stability under predictive perturbations on a common exploration-set family. Furthermore, we propose \agentcbo{}, a graph-belief routing framework that incorporates LLM causal priors through guarded proposal and commit mechanisms. Specifically, a \textsc{Language-Prior Agent} combines LLM beliefs with constraint-based discovery to extract a protected causal backbone alongside residual uncertain edges. A \textsc{Calibration Agent} and a \textsc{Residual Repair Agent} then refine only these uncertain structures through score-gated bounded local edits. Finally, a deterministic commit gate evaluates candidate updates using routing-proxy checks before allowing them to influence intervention selection. Across ten benchmarks, \agentcbo{} performs comparably to known-graph CBO, demonstrating that guarded LLM priors can improve fixed-budget CBO performance.
\end{abstract}

\section{Introduction} \label{sec:introduction}

In scientific research, optimizing complex objectives frequently requires carefully planned sequential experiments because evaluating candidate interventions is typically expensive \citep{jones1998efficient, shahriari2016taking, frazier2018tutorial}. Bayesian optimization (BO) provides a principled framework for this setting by using probabilistic surrogate models and sequential acquisition rules to select informative experiments \citep{mockus1975bayesian, jones1998efficient, shahriari2016taking, frazier2018tutorial}. When causal structure is available, causal Bayesian optimization (CBO) can further improve sample efficiency by exploiting the graph to restrict interventions and initialize graph-informed surrogate models \citep{pearl2009causality, aglietti2020causal}. Concretely, the causal graph routes the optimizer by determining which variables should be considered for intervention, which exploration sets (ESs) are admissible, and which \emph{do}-calculus priors inform the corresponding surrogate models, ultimately shaping the acquisition policy used at each CBO update.

A long-standing challenge in CBO is that the known-graph assumption is rarely realistic in scientific investigations. Recent unknown-graph CBO methods address this limitation by coupling causal optimization with structure uncertainty. Causal Entropy Optimization models uncertainty over causal graphs within the surrogate and acquisition process to trade off structure learning and causal-effect optimization \citep{branchini2023causal}. Graph Agnostic Causal Bayesian Optimization maintains plausible graph models and selects interventions optimistically so that causal discovery is pursued only when it serves reward maximization \citep{mukherjee2024graph}. CBO with Unknown Graphs improves scalability by focusing posterior inference on the direct causal parents of the target variable \citep{durand2025causal}. While these methods weaken the need for expert-specified graphs, a remaining limitation is that graph relevance is still expressed through graph hypotheses, graph posteriors, or structurally restricted target-local summaries. This results in an optimization objective that is too tightly coupled with causal discovery, often demanding significant sample budgets to resolve structural uncertainties that might be irrelevant to the downstream choices of the optimizer.

\begin{figure}[t]
\centering
\begin{tikzpicture}[
  every node/.style={font=\small},
  obs/.style={draw=black!70, line width=0.7pt, circle,
              minimum size=7mm, inner sep=0pt, fill=white},
  cov/.style={draw=black!70, line width=0.7pt, circle,
              minimum size=7mm, inner sep=0pt, fill=gray!18},
  tgt/.style={draw=black!70, line width=0.8pt, circle,
              minimum size=7mm, inner sep=0pt, fill=cyan!25},
  edge/.style={-{Stealth[length=3.0mm,width=2.5mm]},
               line width=0.8pt, draw=black!75},
  cedge/.style={-{Stealth[length=3.0mm,width=2.5mm]},
                line width=1.0pt, draw=red!65!black},
  panel/.style={font=\bfseries\small, inner sep=2pt},
  routing/.style={fill=gray!10, rounded corners=3pt},
]

\begin{scope}
  \node[panel] at (0, 2.10) {$\mathcal{G}_1$};
  \begin{pgfonlayer}{background}
    \fill[routing] (-1.45, 1.80) rectangle (1.45, -1.10);
  \end{pgfonlayer}
  \node[cov] (D1) at ( 0,    1.45) {$D$};
  \node[obs] (A1) at (-1.0,  0.50) {$A$};
  \node[obs] (B1) at ( 1.0,  0.50) {$B$};
  \node[tgt] (Y1) at ( 0,   -0.70) {$Y$};
  \node[obs] (E1) at ( 1.85,-0.70) {$E$};
  \draw[edge] (D1) -- (A1);
  \draw[edge] (A1) -- (Y1);
  \draw[edge] (B1) -- (Y1);
  \draw[edge] (A1) -- (E1);
\end{scope}

\node[font=\large] at (2.85, 0.4) {$\sim_{\mathrm{R}}$};

\begin{scope}[shift={(4.85, 0)}]
  \node[panel] at (0, 2.10) {$\mathcal{G}_2$};
  \begin{pgfonlayer}{background}
    \fill[routing] (-1.45, 1.80) rectangle (1.45, -1.10);
  \end{pgfonlayer}
  \node[cov] (D2) at ( 0,    1.45) {$D$};
  \node[obs] (A2) at (-1.0,  0.50) {$A$};
  \node[obs] (B2) at ( 1.0,  0.50) {$B$};
  \node[tgt] (Y2) at ( 0,   -0.70) {$Y$};
  \node[obs] (E2) at ( 1.85,-0.70) {$E$};
  \draw[edge] (D2) -- (A2);
  \draw[edge] (A2) -- (Y2);
  \draw[edge] (B2) -- (Y2);
  \draw[edge] (A2) -- (E2);
  \draw[edge] (B2) -- (E2);
\end{scope}

\node[font=\large] at (7.70, 0.4) {$\not\sim_{\mathrm{R}}$};

\begin{scope}[shift={(9.70, 0)}]
  \node[panel] at (0, 2.10) {$\mathcal{G}_3$};
  \begin{pgfonlayer}{background}
    \fill[routing] (-1.45, 1.80) rectangle (1.45, -1.10);
  \end{pgfonlayer}
  \node[cov] (D3) at ( 0,    1.45) {$D$};
  \node[obs] (A3) at (-1.0,  0.50) {$A$};
  \node[obs] (B3) at ( 1.0,  0.50) {$B$};
  \node[tgt] (Y3) at ( 0,   -0.70) {$Y$};
  \node[obs] (E3) at ( 1.85,-0.70) {$E$};
  \draw[edge] (D3) -- (A3);
  \draw[edge] (A3) -- (Y3);
  \draw[edge] (B3) -- (Y3);
  \draw[edge] (A3) -- (E3);
  \draw[cedge] (D3) -- (B3);
\end{scope}

\end{tikzpicture}

\caption{\textbf{Routing-equivalence vs.\ routing-critical edits.} All three DAGs share manipulative ancestors $\{A,B\}$ of $Y$; $D$ (gray) is non-manipulable and $E$ is a non-ancestor. The schematic assumes matched fitting inputs and unchanged do-priors for the outside edit $B\!\to\!E$, giving $\mathcal{G}_1\sim_{\mathrm{R}}\mathcal{G}_2$. The inside edit $D\!\to\!B$ (red) is routing-critical if $d_\mu>\delta_\mu$, as assumed for the displayed relation (Defs.~\ref{def:routing_equiv}--\ref{def:routing_critical}).}
\label{fig:routing_eq}
\end{figure}
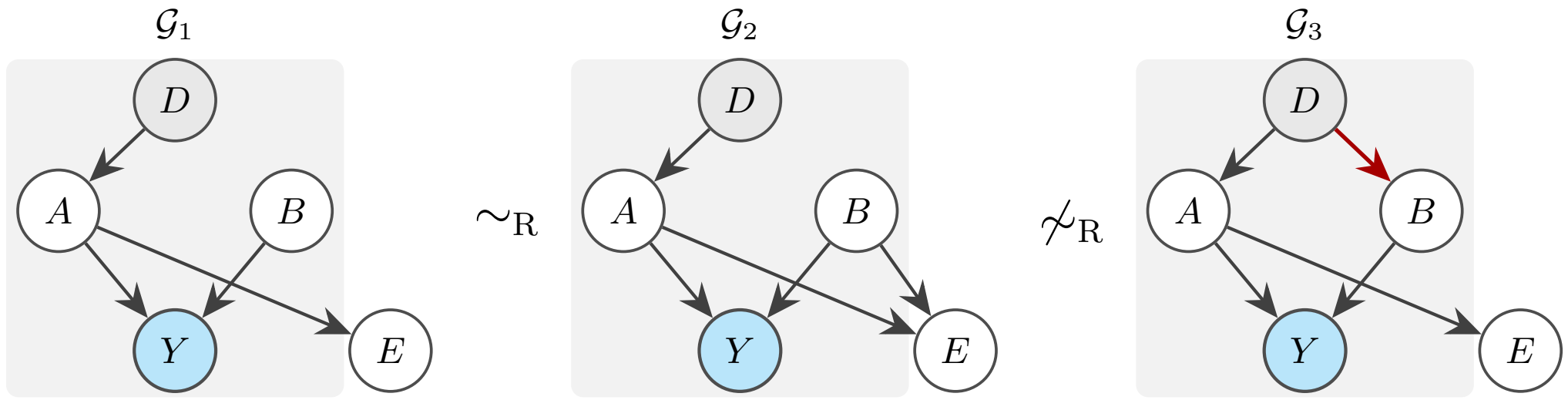

Drawing insights from causal-bandit research, partial structural information has been shown to suffice for no-regret learning even under hidden confounders \citep{elahi2024partial}. A similar principle applies to our fixed-budget Bayesian optimization setting. What matters to the optimizer is not the exact recovery of the entire graph topology, but rather the optimizer-facing routing objects induced by the graph, such as the runtime ES family and the graph-derived \emph{do}-prior objects. This insight leads to the notions of \emph{routing-equivalence} and \emph{routing-critical edits}. At a fixed update, two graphs are routing-equivalent if they induce the same runtime ES family and the same graph-derived prior objects. A local edit is routing-critical when it changes the ES family or moves the prior mean beyond a specified tolerance. As Figure~\ref{fig:routing_eq} illustrates, an edit can change the DAG without changing its routing inputs; at matched state and coupled fitting/selector randomness, such an edit preserves the next intervention. An edit within the target-relevant ancestral structure can instead alter admissible ESs or graph-derived priors. Therefore, in latent-confounded settings, the maintained graph should be interpreted as an observed-variable routing proxy rather than a strictly identifiable causal DAG.

Based on this observation, we reframe unknown-graph CBO as \emph{causal belief routing} rather than full causal-graph recovery. In this paper, we propose \agentcbo{}, a guarded graph-belief routing framework that treats large language model (LLM) causal outputs as structured but uncertain causal priors. \agentcbo{} does not treat the LLM as an autonomous optimizer or an unrestricted causal discovery oracle. Instead, it enforces a strict separation of roles where language models provide structured but uncertain causal beliefs while the BO loop manages intervention selection, surrogate updating, and acquisition. The graph-update stack follows a guarded \emph{Propose, Calibrate, Repair, and Commit} architecture. The \textsc{Language-Prior Agent} elicits pairwise causal beliefs and combines them with constraint-based discovery to obtain a Peter-Clark(PC) Algorithm\citep{spirtes2000causation}-anchored point DAG, a protected backbone, and residual uncertain pairs. The \textsc{Calibration Agent} and \textsc{Residual Repair Agent} then operate only on residual structures through score-gated refinement and bounded local edits. Candidate updates are accepted through a deterministic \textsc{Commit Gate} based on confidence and inexpensive routing-proxy checks. When a graph is committed, runtime ESs and graph-derived \emph{do}-prior objects are recomputed, and archived interventions are re-injected whenever compatible ESs reappear. Interventional outcomes update the per-ES BO surrogate and archive instead of being fed into the LLM-based proposal mechanism.

\textbf{In summary, our contributions are as follows.}
(1) We propose \agentcbo{}, an LLM-assisted CBO framework whose \emph{Propose, Calibrate, Repair, and Commit} stack admits graph beliefs through PC anchoring, score-gated calibration, bounded repair, and a deterministic commit gate, while retaining CBO's intervention authority. (2) We formalize \emph{causal belief routing} through the ES and do-prior interface, giving sufficient conditions for fixed-update decision preservation, a quantitative CEI stability bound, and an acquisition-margin condition for retaining the selected ES. (3) Across ten benchmarks, \agentcbo{} is generally competitive with known-graph CBO, with its largest refinement gain on CompleteGraph. Matched PC-only ablations, exploration-set controls, route replay, and a controlled corruption study provide evidence for prior utility, finite-budget routing effects, and update-time protection. ALARM evaluates semantic-prior utility at 37 nodes, while ordered ER studies demonstrate operation and quantify computational cost through 50 nodes under a supplied ordering prior.

\section{Background and Problem Statement}
\label{sec:background}

\subsection{Structural Causal Models and Interventions}

A system is described by a structural causal model (SCM) $\mathcal{M} = \langle \mathcal{G}, \mathbf{V}, \mathbf{F}, P(\boldsymbol{\omega}) \rangle$, where $\mathcal{G} = (\mathbf{V}, \mathbf{E})$ is a directed acyclic graph (DAG), $\mathbf{V} = \{X_0, \ldots, X_{m-1}, Y\}$ is a set of endogenous observed random variables with target variable $Y = X_m$, $\mathbf{F} = \{f_i\}_{i=0}^{m}$ is a set of unknown structural equations, and $P(\boldsymbol{\omega})$ is a joint distribution over noise variables. Each variable is generated according to $X_i = f_i(\mathbf{Z}_i) + \omega_i$, where $\mathbf{Z}_i = \{X_j : j \in \mathrm{pa}(i)\}$ denotes the parents of $X_i$ in $\mathcal{G}$ and $\omega_i$ is zero-mean noise. The set of \textit{manipulative variables} $\mathbf{M} \subseteq \mathbf{V} \setminus \{Y\}$, on which interventions are possible, is distinguished from the set of non-manipulative variables $\mathbf{C} = \mathbf{V} \setminus \mathbf{M} \setminus \{Y\}$. A \textit{hard intervention} $\mathrm{do}(\mathbf{X}_S = \mathbf{x}_S)$ on a subset $S \subseteq \mathbf{M}$ removes all incoming edges to the variables in $S$ and fixes their values to $\mathbf{x}_S$, inducing the interventional distribution $P(Y \mid \mathrm{do}(\mathbf{X}_S = \mathbf{x}_S))$. The \textit{causal effect} of this intervention on the target is $\mu_S(\mathbf{x}_S) = \mathbb{E}[Y \mid \mathrm{do}(\mathbf{X}_S = \mathbf{x}_S)]$, which can be computed from the SCM when the relevant interventional distribution is identifiable, for example via truncated factorization or do-calculus adjustment formulas.

We allow the data-generating SCM to contain latent variables $\mathbf{U}$, in which case the full causal graph is a DAG over $\mathbf{V} \cup \mathbf{U}$ and the noise-independence assumption $P(\boldsymbol{\omega})$ applies to the joint $(\mathbf{V}, \mathbf{U})$ system; after marginalizing $\mathbf{U}$, observed residuals on $\mathbf{V}$ may be statistically dependent and the observed-variable graph is in general a latent projection rather than an ordinary DAG. \agentcbo{} maintains an observed-variable graph proxy $\hat{\mathcal{G}}$ over $\mathbf{V}$, used to construct the routing channels (exploration sets and do-priors) of Section~\ref{sec:method_routing}: in fully observed benchmarks $\mathbf{U} = \emptyset$ and $\hat{\mathcal{G}}$ targets the true DAG, while in latent-confounded benchmarks such as \textsc{CompleteGraph} (Appendix~\ref{app:benchmarks}), $\hat{\mathcal{G}}$ functions as a routing proxy used by the optimizer rather than an identifiable causal graph.

\subsection{Causal Bayesian Optimization}

CBO aims to minimize the causal effect by jointly selecting the variables for intervention and the corresponding values to assign: $(S^{\star}, \mathbf{x}^{\star}_{S^{\star}}) = \argmin_{S \in \mathcal{ES},\, \mathbf{x}_S \in \mathcal{D}(S)} \mu_S(\mathbf{x}_S)$, where $\mathcal{ES}$ denotes the exploration-set (ES) family, a collection of causally motivated subsets of the manipulative variables, and $\mathcal{D}(S)$ is the interventional domain for $S$. In standard CBO, the ES family is derived from the causal graph. Each $S \in \mathcal{ES}$ is a subset of the manipulative ancestors of $Y$ in $\mathcal{G}$, and the possibly-optimal minimal intervention sets (POMIS) provide the theoretically minimal collection~\citep{lee2018structural}. CBO maintains a separate Gaussian process (GP) surrogate $\hat{f}_S \sim \mathcal{GP}(\mu_S^{\mathrm{prior}}, k_S)$ for each ES $S \in \mathcal{ES}$. The prior mean $\mu_S^{\mathrm{prior}}$ is informed by do-calculus propagation through a network of per-node conditional GPs fitted to observational data. At each iteration, CBO selects the ES and intervention values that maximize the cost-normalized causal expected improvement (CEI):
\begin{equation}
\label{eq:cei}
(S_t, \mathbf{x}_t) = \argmax_{S \in \mathcal{ES},\; \mathbf{x}_S \in \mathcal{D}(S)}\; \frac{\mathrm{EI}_S(\mathbf{x}_S)}{\mathrm{cost}(S, \mathbf{x}_S)}.
\end{equation}
Between interventions, an $\epsilon$-greedy schedule is used to collect additional observational data, update the conditional GP network, and refine the do-calculus prior.

\paragraph{Standard known-graph setting.}
The standard CBO framework assumes that the causal DAG $\mathcal{G}$ is known. This assumption is also used by model-based extensions, such as model-based CBO (MCBO) and adversarial CBO. This work relaxes the known-graph setting. Rather than treating the resulting problem solely as a sequential task of causal discovery followed by optimization, the causal graph $\mathcal{G}$ is considered as the input to a deterministic mapping pipeline for the optimizer: $\mathcal{G} \mapsto \mathcal{ES}(\mathcal{G}) \mapsto \{\mu_S^{\mathrm{prior}}(\mathcal{G}), \sigma_S^{\mathrm{prior}}(\mathcal{G})\}_{S\in\mathcal{ES}(\mathcal{G})} \mapsto \Pi_{\mathcal{G}}$, where $\Pi_{\mathcal{G}}$ denotes the CEI-driven intervention policy of Eq.~\eqref{eq:cei}. In this routing-policy framework, the optimizer interacts with the graph through the induced ESs and the do-calculus prior objects used by the surrogate, rather than through the full topology directly. As a result, two graphs that yield identical outputs along this mapping are indistinguishable to the BO policy via these routing channels, even if they differ in structural details that do not affect target-relevant intervention routes. This perspective motivates the evaluation of estimated graphs by the intervention routes they induce, in addition to structural recovery metrics.

\section{Related Work} \label{sec:related}

\subsection{CBO under Unknown Graphs}
CBO~\citep{aglietti2020causal} reduces intervention costs by leveraging a known causal graph to define exploration sets and do-calculus priors. Subsequent extensions have enhanced CBO in adversarial~\citep{sussex2024adversarial}, constrained~\citep{aglietti2023constrained}, and multi-objective~\citep{bhatija2025multiobjective} contexts; however, most approaches assume the causal graph is known in advance. Relaxing this assumption remains challenging. Existing unknown-graph CBO methods differ along four axes: the structural object maintained, the evidence updating it, the update schedule, and how it couples to the BO loop (Table~\ref{tab:method_positioning}). CEO~\citep{branchini2023causal} maintains a posterior over DAGs updated by joint observational and interventional likelihood, with information-theoretic acquisition. GACBO~\citep{mukherjee2024graph} maintains plausible (graph, function, action) triples updated by Bayesian scores and GP confidence sets, with UCB selection. CBO-U~\citep{durand2025causal} maintains a posterior over the target's direct parents updated by parent-set CI tests, with a parent-aware surrogate. POMIS characterizes possibly-optimal minimal intervention sets for a given graph~\citep{lee2018structural}, while \citet{elahi2024partial} identify partial structure sufficient for no-regret causal-bandit learning. \citet{park2025structural} extend intervention-set characterization to Markov-equivalence classes represented by partial ancestral graphs. CEO also illustrates optimization-equivalent graphs with equal causal-effect prior means and covariances~\citep{branchini2023causal}. Our analysis jointly tracks the graph-dependent RuntimeES and rebuilt predictive interface, establishing conditional fixed-update selector preservation and CEI stability.

\begin{table}[h]
\centering
\caption{Positioning of \agentcbo{} relative to unknown-graph CBO baselines along four dimensions.}
\label{tab:method_positioning}
\footnotesize
\setlength{\tabcolsep}{4pt}
\renewcommand{\arraystretch}{1.15}
\resizebox{\linewidth}{!}{%
\begin{tabular}{@{}lcccc@{}}
\toprule
Method & Structural scope & Evidence & Belief update & BO coupling \\
\midrule
Dynamic PC-only & DAG & CI tests & Periodic & Guarded ESs and priors \\
CEO~\citep{branchini2023causal}   & DAG posterior   & Obs+int likelihood     & Online           & Entropy search \\
GACBO~\citep{mukherjee2024graph} & Plausible DAGs  & Bayes scores + GP CS   & Online           & UCB over $(G,f,a)$ \\
CBO-U~\citep{durand2025causal} & Parents of $Y$  & Parent-set CI          & Online           & Parent-aware surrogate \\
\midrule
\agentcbo{} (ours) & Target-routed DAG & CI + LLM prior, guarded & Periodic, gated & Routing-aware ESs and priors \\
\bottomrule
\end{tabular}%
}
\end{table}

\subsection{LLMs in Causal Discovery and Scientific Optimization}
Recent evaluations reveal that while LLMs excel at pairwise causal judgments, they struggle with full graph recovery and often conflate correlation with causation~\citep{kiciman2024causal,jin2024can,chen2024calm}. A robust mitigation strategy integrates LLM knowledge as soft priors into traditional discovery algorithms, allowing statistical data to override inconsistent LLM beliefs~\citep{long2023imperfect}. LLM-assisted BO uses language models for surrogate modeling and candidate generation~\citep{liu2024large,cisse2025bora,agarwal2025bopro}, while other approaches use them as direct optimizers~\citep{yang2024opro,roohani2025bio}. LGBO~\citep{yuan2026unleashing} incorporates iterative LLM preferences through region-lifted GP mean shifts. LABO~\citep{chen2026labo} combines LLM predictions and experimental observations in a multi-fidelity surrogate with discrepancy-based evaluation gating. \agentcbo{} uses LLM guidance in graph construction and guarded commitment, linking the committed graph to RuntimeES and do-priors while CBO selects interventions.

\section{Causal Belief Routing}
\label{sec:method}

\begin{figure}[t]
\centering
\includegraphics[width=\linewidth]{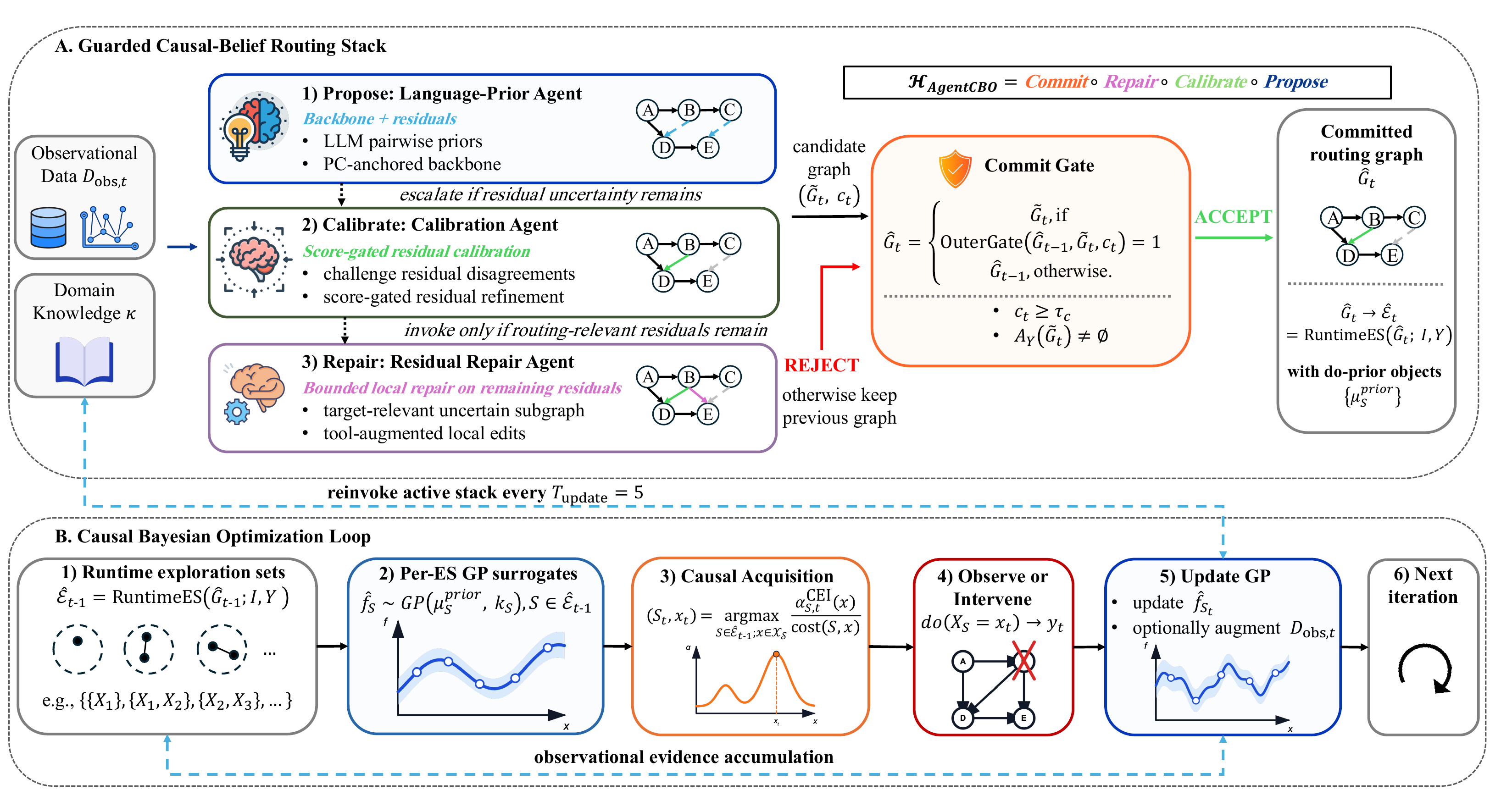}
\caption{
\textbf{\agentcbo{} guarded routing pipeline.}
The \prioragent{} proposes a causal backbone and routing-relevant residual pairs, the \calibagent{} imports score-improving statistical calibrations, and the \repairagent{} applies bounded local repair. The deterministic \commitgate{} compares the stacked candidate \(\tilde G_t\) with the previous committed graph \(\hat G_{t-1}\) and commits the graph used by CBO. The committed graph induces RuntimeES and do-priors for cost-normalized CEI. The outer loop either observes a fresh batch or executes an intervention; interventions update the surrogate and archive, not graph discovery by default.
}
\label{fig:framework}
\end{figure}

\subsection{Problem Setting}
\label{sec:method_setup}

Given an observational dataset $\mathcal{D}_{\mathrm{obs}} = \{\mathbf{v}^{(i)}\}_{i=1}^{n_{\mathrm{obs}}}$ drawn from the observed marginal of an unknown SCM $\mathcal{M}$ whose full graph $\mathcal{G}^{\star}$ may include latent variables (Section~\ref{sec:background}), with manipulative subset $\mathbf{M} \subseteq \mathbf{V}$, \agentcbo{} maintains an estimated graph $\hat{\mathcal{G}}_t$ at each round $t$ and selects interventions with CEI as defined in Eq.~\eqref{eq:cei}. In this section, we write $\mathcal{I}:=\mathbf{M}$ for the intervention-eligible variables to avoid overloading $\mathcal{M}$. Let
$
\mathcal{A}_Y(\mathcal{G}) := \mathrm{anc}_{\mathcal{G}}(Y)\cap\mathcal{I}
$
denote the manipulative ancestors of $Y$. While the ideal CBO construction utilizes $\mathcal{ES}_{\mathrm{ideal}}(\mathcal{G}) := 2^{\mathcal{A}_Y(\mathcal{G})}\setminus\{\emptyset\}$, the implemented optimizer instead employs a runtime constructor
\begin{equation}
\widehat{\mathcal{E}}(\mathcal{G})=\RuntimeES(\mathcal{G};\mathcal{I},Y).
\label{eq:runtime_es}
\end{equation}
This constructor retains singleton manipulative interventions, truncates excessively large families, and defaults to a conservative strategy when no manipulative ancestor is identified. During graph-update rounds, the active stack outputs a candidate graph $\tilde{\mathcal{G}}_t$, a confidence scalar $c_t$, and metadata for the outer gate. Subsequently, the outer loop integrates the committed graph with the conditional GP network fitted on $\mathcal{D}_{\mathrm{obs}}$ to compute the do-calculus prior objects required by CEI.

\subsection{Routing-Equivalence and Routing-Critical Edits}
\label{sec:method_routing}

The routing-policy perspective described in Section~\ref{sec:background} treats the graph as input to the runtime ES constructor, the graph-derived do-prior objects, and ultimately the CEI policy. Consequently, the appropriate unit of analysis is not an individual edge, but rather the equivalence class of graphs that yield identical optimizer-facing routing inputs.

\begin{definition}[Routing-equivalence, fixed update]
\label{def:routing_equiv}
At a fixed CBO update with matched non-routing inputs, two graphs $\mathcal{G}_1,\mathcal{G}_2$ are \textbf{routing-equivalent}, written $\mathcal{G}_1 \sim_{\mathrm{R}} \mathcal{G}_2$, if they induce the same labeled runtime ES family, $\widehat{\mathcal{E}}(\mathcal{G}_1)=\widehat{\mathcal{E}}(\mathcal{G}_2)$, and pointwise-equal graph-derived do-prior objects on every admissible training or query input used by the surrogate builder.
\end{definition}

At matched inputs, labeled graph equality implies routing-equivalence, whereas Markov equivalence is generally not comparable to it. Markov-equivalent graphs can induce different intervention routes, and routing-equivalent graphs can differ away from those routes.

\begin{definition}[Routing-critical edit]
\label{def:routing_critical}
For a local graph edit $\Delta$, let $\mathcal{G}\oplus\Delta$ denote the edited graph. Given a tolerance $\delta_\mu\ge0$, the edit is \textbf{routing-critical} if it changes the runtime ES family, $\widehat{\mathcal{E}}(\mathcal{G})\neq\widehat{\mathcal{E}}(\mathcal{G}\oplus\Delta)$, or if the do-prior mean distance satisfies $d_\mu(\mathcal{G},\mathcal{G}\oplus\Delta)>\delta_\mu$.
\end{definition}

The outer gate described below does not compute this definition exactly. Instead, it uses manipulative target ancestry and normalized structural Hamming distance ($n\mathrm{SHD}$) on the ancestor subgraph as computationally efficient proxies for routing-criticality. The complete distance-based definition and the gate predicate are provided in Appendix~\ref{app:engines}.

\paragraph{Fixed-update guarantees.}
Hold fixed the data, archive, domains, costs, and fitting/selection rules \emph{before} rebuilding route-conditioned surrogates. All graph dependence of the common builder and selector must enter through the labeled ES family and graph-derived prior objects. Fitted predictions are outputs, not fixed inputs. Appendix~\ref{app:routing_theory} gives the full assumptions and proofs.

\begin{lemma}[Exact selector sufficiency]
\label{lem:route_exact}
Under the matched-state conditions of Appendix~\ref{app:route_assumptions}, routing-equivalent graphs give the same conditional next-intervention distribution when the surrogate builder and numerical selector use common graph-independent randomness and tie-breaking.
\end{lemma}

\begin{proposition}[Approximate CEI stability]
\label{prop:cei_stability}
Compare two rebuilt Gaussian predictors on the same finite nonempty labeled ES family and nonempty domains, with the same cost $c(a)\ge c_{\min}>0$ for action $a=(S,\mathbf{x})$. Write $g=b-m$ and $\widetilde g=\widetilde b-\widetilde m$ for minimization improvement means, where $b$ is the EI incumbent, and $s,\widetilde s\ge0$ for predictive standard deviations. Let $\alpha(a)=\mathbb E[(g(a)+s(a)U)_+]/c(a)$ for $U\sim\mathcal N(0,1)$, and define $\widetilde\alpha$ analogously. If $\|g-\widetilde g\|_\infty\le\epsilon_g$ and $\|s-\widetilde s\|_\infty\le\epsilon_s$, with finite bounds, then
\begin{equation}
\|\alpha-\widetilde\alpha\|_\infty\le
\eta:=\frac{\epsilon_g+\epsilon_s/\sqrt{2\pi}}{c_{\min}}.
\label{eq:cei_stability}
\end{equation}
If $\widetilde\alpha(\widetilde a)\ge\sup_a\widetilde\alpha(a)-\zeta$ for finite $\zeta\ge0$, then $\sup_a\alpha(a)-\alpha(\widetilde a)\le2\eta+\zeta$. If the reference-optimal ES has a peak-acquisition margin $\Gamma>2\eta+\zeta$, its identity is preserved (Corollary~\ref{cor:es_margin}).
\end{proposition}

The proof uses the Gaussian-EI sensitivities $|\partial_g\mathrm{EI}|\le1$ and $|\partial_s\mathrm{EI}|\le1/\sqrt{2\pi}$, followed by an approximate-maximizer comparison. These results link the joint ES--surrogate interface to intervention selection without requiring full graph recovery. The guarantees apply at a matched fixed update and require control of both predictive means and uncertainties. The implemented proxy gate is evaluated separately in Appendix~\ref{app:gate_stress}.

\subsection{\agentcbo{}: A Guarded LLM-Agent Stack} \label{sec:method_stack}

\paragraph{Setup.}
\agentcbo{} instantiates the routing view with a guarded graph-update stack. At update rounds, the \prioragent{} proposes a point DAG, frozen backbone, and residual candidate pairs; the \calibagent{} and \repairagent{} may modify only those residual pairs, and only when the projected edit improves the internal structural score. The deterministic \commitgate{} then decides whether the candidate graph replaces the graph currently routing CBO. The reported variants activate progressively more stages: \agentcboprior{} uses proposal only, \agentcbocalib{} adds calibration, and \agentcbofull{} adds residual repair. Interventional observations update the BO surrogate and archive, not the LLM-based graph proposal. The full LLM input--output contract for each agent is specified in Appendix~\ref{app:llm_interface}. We use ``belief'' operationally, not as a Bayesian posterior over DAGs: the router maintains a committed graph $\hat{\mathcal{G}}_t$, a Prior-Agent proposal $\mathcal{G}^{\mathrm{LP}}_t$, a frozen backbone $\mathcal{B}_t$, residual pairs $\mathcal{R}_t$, an internal structural score $s(\mathcal{G})$, the runtime ES family $\widehat{\mathcal{E}}_t$, and a cross-graph archive $\mathcal{H}_t$.

\paragraph{Propose stage.}
The \prioragent{} combines pairwise LLM causal beliefs with PC anchoring, CPDAG-to-DAG extension, guarded edge supplementation, and confidence scoring. It also constructs a frozen backbone and residual pairs:
\begin{equation}
\left(\mathcal{G}^{\mathrm{LP}}_t,\mathcal{B}_t,\mathcal{R}_t,c^{\mathrm{LP}}_t\right)
=
\Propose_{\mathrm{LP}}(\mathcal{D}_{\mathrm{obs},t},\kappa),
\qquad
\mathcal{R}_t =
\left(\mathcal{R}^{\mathrm{low}}_t \cup \mathcal{R}^{\mathrm{conflict}}_t \cup \mathcal{R}^{\mathrm{target}}_t\right)\setminus\mathcal{B}_t .
\label{eq:lp_residual}
\end{equation}
The complete LP details are given in Algorithm~\ref{alg:lp} in Appendix~\ref{app:engines}.

\paragraph{Refine stages.}
Starting from $\mathcal{G}^0_t=\mathcal{G}^{\mathrm{LP}}_t$, the calibration and repair stages share one guarded refinement interface. For stages $k\in\{1,2\}$, let $a_1=\mathrm{CC}$ and $a_2=\mathrm{RA}$. The stage proposes a challenger graph $\mathcal{G}^{a_k}_t$, and the stack imports only edits that lie in the active residual set, preserve the frozen backbone, obey the stage edit budget, and keep the graph acyclic:
\begin{equation}
\bar{\mathcal{G}}^k_t =
\Project\!\left(
\mathcal{G}^{a_k}_t;\,
\mathcal{G}^{k-1}_t,\mathcal{R}^k_t,\mathcal{B}_t,m_k
\right).
\label{eq:guarded_project}
\end{equation}
The projected challenger is accepted only when it improves the internal structural score:
\begin{equation}
\mathcal{G}^k_t =
\begin{cases}
\bar{\mathcal{G}}^k_t, & s(\bar{\mathcal{G}}^k_t) < s(\mathcal{G}^{k-1}_t)-\Delta_k,\\ \mathcal{G}^{k-1}_t, & \text{otherwise}.
\end{cases}
\label{eq:guarded_accept}
\end{equation}
Here $m_1=\infty$ for the \calibagent{} and $m_2=m_{\max}$ for the \repairagent{}. We use $\Delta_1=\Delta_2=0$ and $m_{\max}=5$; acceptance-rate diagnostics and threshold rationale are reported in Appendix~\ref{app:thresholds}. If no projected edit improves the score, the stage is a no-op, and the stack returns the single candidate $\tilde{\mathcal{G}}_t=\mathcal{G}^2_t$ to the \commitgate{}.

\paragraph{Score function.}
The internal score is a statistical structural score, with lower values preferred:
\begin{equation}
s(\mathcal{G}) =
\mathrm{NLL}_{\mathrm{holdout}}(\mathcal{G})
+\lambda_{\mathrm{CI}} r_{\mathrm{CI}}(\mathcal{G})
+\lambda_{\mathrm{sp}}\mathrm{SP}(\mathcal{G}).
\label{eq:guarded_score}
\end{equation}
The three terms are holdout structural-equation negative log-likelihood, a conditional-independence (CI)-consistency penalty over d-separation implications, and a scaled edge-count sparsity penalty.

\paragraph{Guarded design.}
The \prioragent{}'s proposal is \textbf{protected by default}: later stages refine only residual pairs through the guarded interface (Eqs.~\eqref{eq:guarded_project}--\eqref{eq:guarded_accept}), with the proposal preserved when no projected edit improves the score. The full stack therefore reduces to \agentcboprior{} when the prior graph is sufficient, to \agentcbocalib{} when no repair edit is justified, or commits a repaired graph only when both the internal score and the \commitgate{} support the change.

\subsection{Outer Loop, Commit Gate, and Cross-Graph Archival}
\label{sec:method_outer_loop}
\label{sec:method_es}%

The active discovery stack is invoked inside a shared outer loop that interleaves observation, intervention, and graph rediscovery. At each iteration, the loop either collects a fresh observational batch or executes an intervention according to an $\epsilon$-greedy observe-versus-intervene schedule (formalized in Algorithm~\ref{alg:cbo}). Observational rounds augment $\mathcal{D}_{\mathrm{obs}}$, refit the per-node conditional GP network, and refresh the do-priors $\mu_S^{\mathrm{prior}}$; intervention rounds select $(S_t, \mathbf{x}_t)$ by cost-normalized CEI and update the corresponding per-ES GP. Every $T_{\mathrm{update}} = 5$ iterations the selected \agentcboprior{}, \agentcbocalib{}, or \agentcbofull{} stack is re-invoked using the current observational data; in the calibrated and full variants, the \calibagent{} and \repairagent{} are conditional refinement stages and a no-op outcome is part of the algorithm when calibration finds no accepted residual edit or repair finds no routing-critical correction worth committing.

\paragraph{Commit Gate.} The stack returns one candidate $\tilde{\mathcal{G}}_t$; the deterministic \commitgate{} decides whether this candidate replaces the currently committed graph:
\begin{equation}
\hat{\mathcal{G}}_t =
\begin{cases}
\tilde{\mathcal{G}}_t, & \OuterGate(\hat{\mathcal{G}}_{t-1},\tilde{\mathcal{G}}_t,c_t)=1,\\ \hat{\mathcal{G}}_{t-1}, & \text{otherwise}.
\end{cases}
\label{eq:outer_commit}
\end{equation}
The \commitgate{} is a cheap proxy for routing-criticality rather than a direct RuntimeES-distance optimizer: it requires sufficient confidence, rejects candidates with no manipulative ancestor of $Y$, and raises the confidence bar when the target-ancestor subgraph changes substantially. The full threshold predicate is given in Appendix~\ref{app:engines}, Eq.~\eqref{eq:full_acceptance}.

\paragraph{Cross-graph data archival.}
A practical bottleneck in sequential discovery-optimization is the stranding of interventional data when the accepted point graph changes. When \commitgate{} triggers a graph change, \agentcbo{} archives all observed interventional data $\{(S, \mathbf{x}_S, y)\}$ keyed by their ES signature $S$. Upon any subsequent graph change that re-admits $S$, these archived tuples are re-injected into the per-ES GP. This is specific to our point-graph switching loop: continuous-posterior methods update structural beliefs directly rather than requiring explicit ES re-admission.

\paragraph{Surrogate and acquisition.}
\label{sec:method_surrogate}%
For each active $S \in \widehat{\mathcal{E}}_t$, we maintain a GP $\hat{f}_S$ with the CausalRBF kernel, in which the do-calculus variance of the prior mean contributes an additive term that inflates the uncertainty induced by prior propagation under the accepted graph. The prior mean $\mu_S^{\mathrm{prior}}$ is computed by the outer loop from $\hat{\mathcal{G}}$ and the per-node conditional GPs fitted on $\mathcal{D}_{\mathrm{obs}}$. We follow the standard CBO $\epsilon$-greedy observe-versus-intervene schedule, and intervention rounds use
\begin{equation}
(S_t,\mathbf{x}_t)
=
\argmax_{S\in\widehat{\mathcal{E}}(\hat{\mathcal{G}}_{t-1}),\,\mathbf{x}\in\mathcal{D}(S)}
\frac{\mathrm{EI}_{S,t}(\mathbf{x})}{\mathrm{cost}(S, \mathbf{x})}.
\label{eq:routed_action}
\end{equation}
Across all reported benchmarks we set $\mathrm{cost}(S, \mathbf{x}) = |S|$, penalizing the cardinality of the intervention set; this matches the cardinality-cost convention used by standard CBO~\citep{aglietti2020causal}.

\begin{algorithm}[h]
\caption{AgentCBO outer loop with active graph-update stack.}
\label{alg:cbo}
\begin{algorithmic}[1]
\Require Budget $T$; rediscovery period $T_{\mathrm{update}}$; observational data $\mathcal{D}_{\mathrm{obs}}$; domain context $\kappa$.
\Ensure Best routed intervention $(S^\star, \mathbf{x}^\star)$ with target value $Y^\star$.
\State $(\hat{\mathcal{G}}_0, \mathcal{B}_0, \mathcal{R}_0, c_0) \gets \StackUpdate(\mathcal{D}_{\mathrm{obs}}, \mathbf{V}, \kappa)$ \Comment{\agentcboprior{} / \agentcbocalib{} / \agentcbofull{}}
\State $\widehat{\mathcal{E}}_0 \gets \RuntimeES(\hat{\mathcal{G}}_0; \mathcal{I}, Y)$;~fit per-ES GPs $\{\hat{f}_S\}$ with CausalRBF kernel.
\For{$t = 1, \ldots, T$}
  \State $(\hat{\mathcal{G}}_t, \widehat{\mathcal{E}}_t) \gets (\hat{\mathcal{G}}_{t-1}, \widehat{\mathcal{E}}_{t-1})$ \Comment{carry forward}
  \State $b_t \sim \mathrm{Bernoulli}(\epsilon_t)$ \Comment{$\epsilon$-greedy schedule}
  \If{$b_t = 1$} \Comment{observe}
    \State Augment $\mathcal{D}_{\mathrm{obs}}$;~refit per-node GPs and do-priors $\{\mu_S^{\mathrm{prior}}\}$.
  \Else \Comment{intervene}
    \State $(S_t, \mathbf{x}_t) \gets \argmax_{S \in \widehat{\mathcal{E}}_{t-1},\, \mathbf{x} \in \mathcal{D}(S)} \mathrm{EI}_{S,t}(\mathbf{x}) / \mathrm{cost}(S, \mathbf{x})$ \Comment{Eq.~\eqref{eq:routed_action}}
    \State $y_t \gets \mathrm{do}(\mathbf{x}_t)$;~update $\hat{f}_{S_t}$.
  \EndIf
  \If{$t \bmod T_{\mathrm{update}} = 0$} \Comment{rediscover}
    \State $(\tilde{\mathcal{G}}_t, \mathcal{B}_t, \mathcal{R}_t, c_t) \gets \StackUpdate(\mathcal{D}_{\mathrm{obs}}, \mathbf{V}, \kappa)$
    \If{$\OuterGate(\hat{\mathcal{G}}_{t-1}, \tilde{\mathcal{G}}_t, c_t) = 1$} \Comment{Eq.~\eqref{eq:full_acceptance}}
      \State $\hat{\mathcal{G}}_t \gets \tilde{\mathcal{G}}_t$;~$\widehat{\mathcal{E}}_t \gets \RuntimeES(\hat{\mathcal{G}}_t; \mathcal{I}, Y)$;~re-fit $\{\hat{f}_S\}$ on archived data.
    \EndIf
  \EndIf
\EndFor
\State \Return $(S^\star, \mathbf{x}^\star, Y^\star)$ from $\argmin_{t \le T,\, b_t = 0}\, y_t$.
\end{algorithmic}
\end{algorithm}

\section{Experiments}
\label{sec:experiments}

\subsection{Experimental Setup}
\label{sec:exp_setup}

\paragraph{Benchmarks and baselines.}
The evaluation is conducted on ten synthetic and semi-synthetic DAGs spanning four categories: small toy graphs (\textsc{Chain-5}, \textsc{ToyGraph}); application-motivated structures (\textsc{PSAGraph}, \textsc{HIVEpidemiology}, \textsc{CCBOHealth}); multi-ancestor graphs (\textsc{PISHCATGraph}, \textsc{PredatorPrey}); and the hidden-confounder benchmark (\textsc{CompleteGraph}), along with standard optimization surfaces (\textsc{Rosenbrock}, \textsc{Ackley}) reformulated as causal problems. The number of observed variables in these graphs ranges from 3 to 11. Appendix~\ref{app:benchmarks} summarizes benchmark details.
The evaluation includes comparisons with graph-free references, \textbf{BO}, \textbf{LLAMBO}, and \textbf{BORA}; a known-graph \textbf{CBO} reference; and unknown-graph CBO baselines, \textbf{CBO-U}, \textbf{Graph-Agnostic CBO (GACBO)}, and \textbf{Causal Entropy Optimization (CEO)}. Within AgentCBO, \agentcboprior{} uses only the proposal layer, \agentcbocalib{} adds statistical calibration of residual disagreements, and \agentcbofull{} adds bounded residual repair. The main matrix reports all three AgentCBO variants; standalone \textbf{CC} and \textbf{RA} are reported only as separate module diagnostics. Additional diagnostics and the dynamic PC-only ablation are in Appendix~\ref{app:full_results}.
Our code is available at \url{https://github.com/Qiyu-Wei/AgentCBO_Official}.

\paragraph{Protocol.}
Each experimental run begins with $n_{\mathrm{obs}}=100$ observational samples and a total optimization budget of $T=50$ outer-loop steps. The result matrix uses 20 random seeds. All LLM-based engines utilize a Qwen3.5-35B-A3B model with vLLM; the serving setups are documented in Appendix~\ref{app:engines}. We report the best objective value $Y^{\star}$ achieved within the budget, along with the mean and standard deviation across seeds. All main experiments use this fixed budget; systematic sweeps over the intervention budget are left to future work.

\subsection{Benchmark Result Matrix and Routing Effects}
\label{sec:exp_main}
\begin{figure}[t]
\centering
\includegraphics[width=\linewidth]{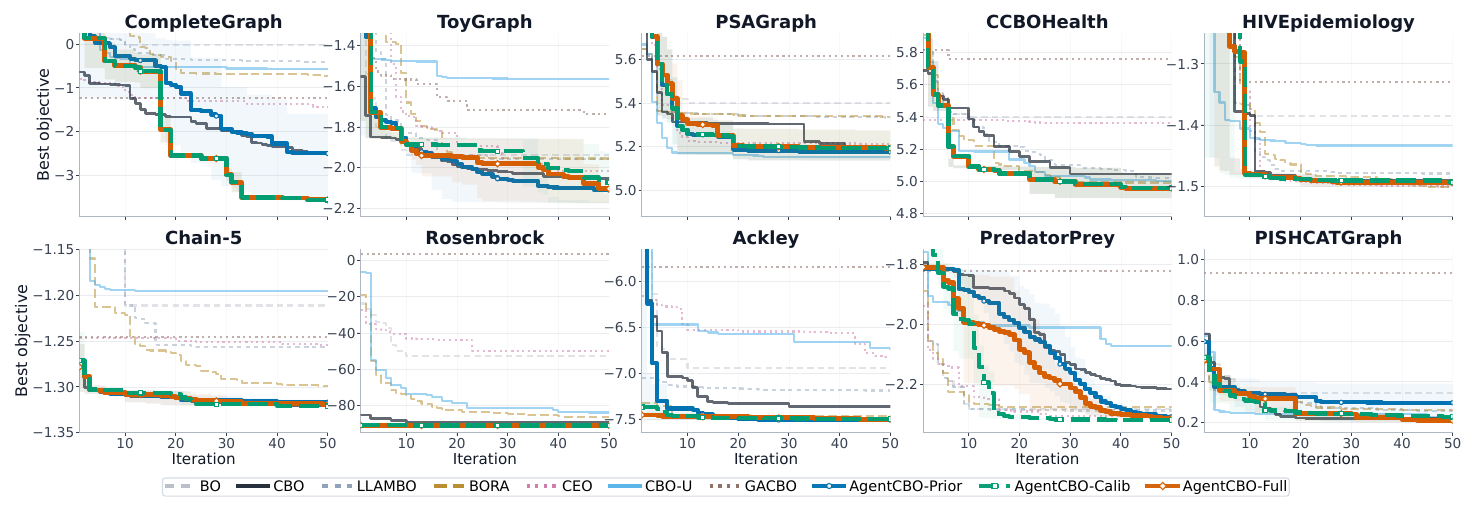}
\caption{\textbf{Best-objective trajectories across the benchmark suite.}
Per-benchmark best-objective curves vs.\ iteration ($T=50$), aggregated
across 20 seeds with shaded variance bands.}
\label{fig:simregret_all}
\end{figure}

Figure~\ref{fig:simregret_all} shows the ten-benchmark trajectories; Table~\ref{tab:main_results} reports their endpoints. The variants are generally competitive with known-graph CBO. Paired tests across benchmark means favor Prior over BO and GACBO after Holm correction across nine comparisons (both adjusted $p=0.018$; Appendix~\ref{app:suite_stats}). Refinement gains vary across the reported benchmark means. Suite-level tests detect no difference between Prior and Calib ($p=0.84$) or Full ($p=0.83$); the CompleteGraph controls below examine the largest observed gain.

\begin{table}[t]
\centering
\caption{%
\textbf{Main result matrix.} Mean $\pm$ standard deviation of best objective $Y^{\star}$ reached within 50 outer-loop steps, aggregated across 20 seeds per benchmark--algorithm pair.
All benchmarks are minimization tasks. \textbf{Bold} marks the best displayed mean among methods other than the known-graph CBO reference baseline; ties at the displayed precision are bolded together.%
}
\label{tab:main_results}
\scriptsize
\setlength{\tabcolsep}{2.5pt}
\renewcommand{\arraystretch}{1.22}
\resizebox{\linewidth}{!}{%
\begin{tabular}{@{}l *{10}{r@{$\pm$}l}@{}}
\toprule
 & \multicolumn{2}{c}{\textit{Reference}}
 & \multicolumn{8}{c}{\textit{CBO baselines}}
 & \multicolumn{4}{c}{\textit{LLM-assisted BO}}
 & \multicolumn{6}{c}{\textit{AgentCBO variants}} \\
\cmidrule(lr){2-3}\cmidrule(lr){4-11}\cmidrule(lr){12-15}\cmidrule(l){16-21}
Benchmark & \multicolumn{2}{c}{BO} & \multicolumn{2}{c}{CBO} & \multicolumn{2}{c}{CEO} & \multicolumn{2}{c}{CBO-U} & \multicolumn{2}{c}{GACBO} & \multicolumn{2}{c}{LLAMBO} & \multicolumn{2}{c}{BORA} & \multicolumn{2}{c}{Prior} & \multicolumn{2}{c}{Calib} & \multicolumn{2}{c}{Full} \\
\midrule
CompleteGraph    & $-0.01$ & $0.35$ & $-2.48$ & $1.14$ & $-1.43$ & $0.55$ & $-0.57$ & $0.16$ & $-1.23$ & $0.12$ & $-0.40$ & $0.35$ & $-0.73$ & $0.04$ & $-2.50$ & $1.21$ & $\mathbf{-3.57}$ & $\mathbf{0.07}$ & $\mathbf{-3.57}$ & $\mathbf{0.07}$ \\
ToyGraph         & $-1.94$ & $0.22$ & $-2.08$ & $0.15$ & $-2.02$ & $0.09$ & $-1.56$ & $0.32$ & $-1.74$ & $0.36$ & $-1.95$ & $0.12$ & $-1.92$ & $0.16$ & $\mathbf{-2.11}$ & $\mathbf{0.17}$ & $-2.02$ & $0.16$ & $-2.10$ & $0.14$ \\
PSAGraph         & $+5.40$ & $0.09$ & $+5.19$ & $0.15$ & $+5.22$ & $0.06$ & $\mathbf{+5.15}$ & $\mathbf{0.02}$ & $+5.62$ & $0.02$ & $+5.34$ & $0.01$ & $+5.34$ & $0.01$ & $+5.18$ & $0.07$ & $+5.19$ & $0.08$ & $+5.19$ & $0.08$ \\
CCBOHealth       & $+5.40$ & $0.14$ & $+5.05$ & $0.21$ & $+5.36$ & $0.11$ & $+5.00$ & $0.06$ & $+5.83$ & $0.19$ & $+5.02$ & $0.12$ & $+4.99$ & $0.14$ & $\mathbf{+4.96}$ & $\mathbf{0.09}$ & $\mathbf{+4.96}$ & $\mathbf{0.09}$ & $\mathbf{+4.96}$ & $\mathbf{0.09}$ \\
HIVEpidemiology  & $-1.39$ & $0.04$ & $-1.49$ & $0.01$ & $\mathbf{-1.50}$ & $\mathbf{0.02}$ & $-1.43$ & $0.08$ & $-1.33$ & $0.41$ & $-1.48$ & $0.01$ & $\mathbf{-1.50}$ & $\mathbf{0.01}$ & $\mathbf{-1.50}$ & $\mathbf{0.01}$ & $-1.49$ & $0.01$ & $-1.49$ & $0.01$ \\
Chain-5          & $-1.21$ & $0.09$ & $-1.32$ & $0.00$ & $-1.25$ & $0.07$ & $-1.20$ & $0.13$ & $-1.25$ & $0.01$ & $-1.26$ & $0.07$ & $-1.30$ & $0.02$ & $\mathbf{-1.32}$ & $\mathbf{0.01}$ & $\mathbf{-1.32}$ & $\mathbf{0.01}$ & $\mathbf{-1.32}$ & $\mathbf{0.01}$ \\
Rosenbrock       & $-52.85$ & $16.57$ & $-89.45$ & $8.23$ & $-49.86$ & $31.37$ & $-83.88$ & $2.93$ & $+3.68$ & $0.10$ & $-90.60$ & $0.49$ & $-86.19$ & $3.52$ & $-90.98$ & $0.05$ & $\mathbf{-90.99}$ & $\mathbf{0.02}$ & $\mathbf{-90.99}$ & $\mathbf{0.02}$ \\
Ackley           & $-6.94$ & $0.36$ & $-7.39$ & $0.02$ & $-6.82$ & $0.41$ & $-6.74$ & $0.54$ & $-5.84$ & $0.04$ & $-7.14$ & $0.21$ & $-7.43$ & $0.02$ & $\mathbf{-7.51}$ & $\mathbf{0.01}$ & $-7.50$ & $0.01$ & $-7.50$ & $0.01$ \\
PredatorPrey     & $-2.28$ & $0.04$ & $-2.21$ & $0.02$ & $-2.31$ & $0.02$ & $-2.07$ & $0.19$ & $-1.82$ & $0.01$ & $-2.29$ & $0.03$ & $-2.28$ & $0.09$ & $-2.31$ & $0.01$ & $\mathbf{-2.32}$ & $\mathbf{0.00}$ & $-2.31$ & $0.01$ \\
PISHCATGraph     & $+0.34$ & $0.05$ & $+0.22$ & $0.01$ & $+65.28$ & $0.11$ & $+0.22$ & $0.04$ & $+0.90$ & $0.01$ & $+0.25$ & $0.03$ & $+0.26$ & $0.02$ & $+0.30$ & $0.10$ & $+0.22$ & $0.09$ & $\mathbf{+0.21}$ & $\mathbf{0.09}$ \\
\bottomrule
\end{tabular}%
}
\end{table}

\paragraph{CompleteGraph mechanism.}
On the hidden-confounder benchmark, Calib and Full improve the reported endpoint from Prior's $-2.50$ to $-3.57$. Separate fixed-graph controls hold cardinality at three sets: the $\{B\},\{D\},\{B,D\}$ family reaches $y\leq-3.0$ on 14/20 seeds, versus 4/20 for random pair-induced families and 0/20 for either alternative pair. In one successful traced run, a step-15 commit contracts RuntimeES from seven to four sets, increasing $\{B,D\}$ selection from $1/11$ to $18/27$ intervention decisions. Replaying this schedule under the known graph recovers about 84\% of that seed's endpoint gap. These complementary controls connect the gain to finite-budget concentration and retained-set composition (Appendix~\ref{app:cg_controls}).

\subsection{Module Diagnostics and Stage Ablations}
\label{sec:exp_ladder}

\paragraph{Module diagnostics.}
\label{sec:exp_engines}
The matched no-LLM arm is \emph{dynamic PC-only AgentCBO}. It replaces only the graph-proposal engine while retaining the same guarded CBO pipeline and settings. Prior improves on eight of ten benchmarks, ties on \textsc{Chain-5}, and trails by $0.02$ on \textsc{PISHCATGraph} (Appendix~\ref{app:pc_only}). Standalone CC/RA diagnostics are separate from the stacked Calib/Full ablations (Table~\ref{tab:diagnostic_panels}). The standalone RA engine uses about $34\times$ LP's tokens (Table~\ref{tab:engine_cost}), measured independently of the stacked variants.

\paragraph{Routing channels and the implemented guard.}
Of 323 graph pairs with identical RuntimeES, 141 (44\%) exceed a fitted prior-mean-distance threshold of $0.05$, demonstrating that the prior channel carries information beyond ES equality (Appendix~\ref{app:routing_diagnostics}). A separate \textsc{PSAGraph} stress test starts from a valid incumbent and disables coverage rescue in all arms. The prescribed corrupted update causes zero endpoint degradation with the gate active; bypassing the gate worsens all ten matched endpoints, with mean paired degradation $+0.1387$ (95\% CI $[+0.0086,+0.2839]$). This tests update-time protection under the specified corruption, alongside the conditional interface theory (Appendix~\ref{app:gate_stress}).

\begin{table}[t]
\centering
\caption{%
\textbf{Module and semantic-robustness diagnostics.} Mean $\pm$ standard
deviation of best $Y^{\star}$ over 20 seeds; \textbf{Bold} marks the best mean among executable-engine diagnostics.%
}
\label{tab:diagnostic_panels}
\label{tab:executable_results}
\label{tab:anon_ablation}
\tiny
\setlength{\tabcolsep}{2.2pt}
\renewcommand{\arraystretch}{1.08}
\resizebox{\linewidth}{!}{%
\begin{tabular}{@{}l r@{$\,\pm\,$}l r@{$\,\pm\,$}l r@{$\,\pm\,$}l r@{$\,\pm\,$}l r@{$\,\pm\,$}l r@{$\,\pm\,$}l@{}}
\toprule
& \multicolumn{6}{c}{Executable-engine diagnostics} & \multicolumn{4}{c}{Semantic-robustness ablations} & \multicolumn{2}{c}{Component ablation} \\
\cmidrule(lr){2-7}\cmidrule(lr){8-11}\cmidrule(l){12-13}
Benchmark & \multicolumn{2}{c}{LP} & \multicolumn{2}{c}{CC} & \multicolumn{2}{c}{RA} & \multicolumn{2}{c}{LP-Anon} & \multicolumn{2}{c}{LP-Shuffled} & \multicolumn{2}{c}{Dynamic PC-only} \\
\midrule
CompleteGraph    & $\mathbf{-2.50}$ & $\mathbf{1.21}$ & $-1.71$ & $1.43$ & $-1.62$ & $1.32$ & $-1.97$ & $1.59$ & $-1.97$ & $1.59$ & $-2.13$ & $1.45$ \\
ToyGraph         & $\mathbf{-2.11}$ & $\mathbf{0.17}$ & $-2.07$ & $0.14$ & $-2.10$ & $0.12$ & $-2.10$ & $0.12$ & $-2.10$ & $0.12$ & $-2.02$ & $0.16$ \\
PSAGraph         & $\mathbf{+5.18}$ & $\mathbf{0.07}$ & $+5.35$ & $0.13$ & $+5.36$ & $0.13$ & $+5.31$ & $0.12$ & $+5.24$ & $0.09$ & $+5.35$ & $0.13$ \\
CCBOHealth       & $+4.96$ & $0.09$ & $\mathbf{+4.94}$ & $\mathbf{0.08}$ & $+4.96$ & $0.09$ & $+4.92$ & $0.06$ & $+4.92$ & $0.06$ & $+4.98$ & $0.09$ \\
HIVEpidemiology  & $\mathbf{-1.50}$ & $\mathbf{0.01}$ & $-1.49$ & $0.01$ & $-1.49$ & $0.01$ & $-1.49$ & $0.01$ & $-1.49$ & $0.01$ & $-1.48$ & $0.01$ \\
Chain-5          & $\mathbf{-1.32}$ & $\mathbf{0.01}$ & $\mathbf{-1.32}$ & $\mathbf{0.01}$ & $\mathbf{-1.32}$ & $\mathbf{0.01}$ & $-1.32$ & $0.01$ & $-1.32$ & $0.01$ & $-1.32$ & $0.01$ \\
Rosenbrock       & $-90.98$ & $0.05$ & $\mathbf{-91.00}$ & $\mathbf{0.00}$ & $-83.10$ & $24.97$ & $-91.00$ & $0.01$ & $-91.00$ & $0.01$ & $-89.80$ & $0.02$ \\
Ackley           & $\mathbf{-7.51}$ & $\mathbf{0.01}$ & $-7.50$ & $0.01$ & $-6.76$ & $0.76$ & $-7.51$ & $0.01$ & $-7.51$ & $0.01$ & $-7.49$ & $0.01$ \\
PredatorPrey     & $\mathbf{-2.31}$ & $\mathbf{0.01}$ & $-2.21$ & $0.22$ & $-2.17$ & $0.23$ & $-2.25$ & $0.14$ & $-2.20$ & $0.19$ & $-2.17$ & $0.22$ \\
PISHCATGraph     & $+0.30$ & $0.10$ & $+0.21$ & $0.00$ & $\mathbf{+0.20}$ & $\mathbf{0.01}$ & $+0.20$ & $0.00$ & $+0.23$ & $0.07$ & $+0.28$ & $0.01$ \\
\bottomrule
\end{tabular}%
}
\end{table}

\paragraph{Semantic robustness.}
\label{sec:exp_anon}
LP-Anon replaces variable names with opaque tokens $V_i$ and omits domain descriptions; LP-Shuffled permutes the variable-to-description mapping. Both match or improve Prior on five of ten benchmarks (Table~\ref{tab:diagnostic_panels}), showing that performance is not driven solely by lexical names, while degradation on selected tasks supports a role for semantic cues. Appendix~\ref{app:llm_interface} details the semantic and observational inputs available to each agent.

\subsection{Larger-Graph Evaluation}
\label{sec:exp_scale}
On a 37-node ALARM-based synthetic SEM, the prior improves over matched PC-only on all ten seeds when coverage rescue is disabled for component isolation (two-sided sign-test $p=0.002$); enabling the default rescue makes the arms tie. On generated ER graphs with 10, 20, and 50 nodes, a disclosed ordering prior yields favorable paired normalized-gap point estimates at 20 and 50 nodes, but the single-size confidence intervals include or touch zero. Median optimization time is lower than PC-only at both larger sizes. These studies evaluate semantic-prior utility on a public topology and operation under supplied structural information, respectively (Appendix~\ref{app:larger_graphs}).

\section{Conclusion}
\label{sec:conclusion}

\agentcbo{} integrates uncertain LLM causal priors into CBO through guarded graph updates while leaving intervention selection to the optimizer. Fixed-update selector preservation and CEI stability explain the joint role of RuntimeES and do-priors. Matched ablations, family controls, single-run route replay, and a corruption study provide complementary evidence for prior utility, finite-budget routing effects, and update-time protection. Larger-graph studies demonstrate operation through 50 nodes under supplied structural information. Together, these results support guarded graph-belief updates as a practical interface between LLM priors and CBO when no trusted graph is available.

\clearpage

\begin{ack}
We thank the anonymous reviewers for their insightful comments and constructive suggestions. We also acknowledge the University of Manchester's CSF3 facility for providing the GPU resources used in this work.
\end{ack}

\bibliographystyle{plainnat}
\bibliography{references}

\clearpage
\appendix

\section{Analysis and Discussion}
\label{sec:analysis}

\agentcbo{} targets graph beliefs that improve intervention selection through the ES--prior interface. The main text gives fixed-update guarantees and empirical analysis; this appendix records their scope and broader-impact considerations.

\subsection{Limitations}
\label{sec:analysis_limitations}

We treat the following as open problems that bound the scope of our claims:

\begin{enumerate}[leftmargin=*,itemsep=2pt,topsep=2pt]
    \item \textbf{Fixed-update scope.} At matched non-routing state and under common surrogate and selection rules, Lemma~\ref{lem:route_exact} and Proposition~\ref{prop:cei_stability} establish fixed-update selector sufficiency and CEI stability. Extending these guarantees to the full adaptive run requires joint analysis of graph updates, archived ES data, and CEI allocation.
    \item \textbf{Prior knowledge and evaluation scope.} Main-suite experiments use one LLM family, with a four-benchmark size comparison (Appendix~\ref{app:llm_size_scaling}). Name obfuscation does not exclude training exposure. ALARM uses a public topology and real labels; ER supplies the generation ordering. These studies evaluate semantic-prior utility and operation under supplied structural information, not general causal reasoning with unknown orientations. The corruption study evaluates the gate's response to one prescribed erroneous update from a valid incumbent.
    \item \textbf{Conditional refinement.} Calib and Full improve selected tasks, notably \textsc{CompleteGraph}. Suite-level tests do not establish consistent gains over Prior (Appendix~\ref{app:suite_stats}). Selecting refinement effort adaptively and measuring its marginal stacked cost remain useful extensions.
\end{enumerate}

\subsection{Broader Impacts}
\label{sec:analysis_broader}

The main positive impact of AgentCBO is procedural: if graph discovery inside the loop can reduce the number of costly wet-lab or simulator interventions needed to reach a good operating point, then practitioners can save both time and experimental budget. The main negative risk is over-trust. In high-stakes domains such as medicine, policy, or biology, an incorrect discovered graph can still steer the optimization toward the wrong intervention family even when the final BO layer is well calibrated. We therefore view AgentCBO as a decision-support tool rather than an autonomous decision-maker, and we recommend human oversight plus benchmark-appropriate validation before any real deployment.

\section{Benchmark Suite}
\label{app:benchmarks}

This appendix specifies the ten benchmarks used in the main experiments. Table~\ref{tab:benchmark_specs} summarizes the key specifications; the flagship \textsc{CompleteGraph} benchmark is detailed in full below; the remaining nine benchmarks reuse the structural-equation models (SEMs) of their original sources, cited per row.

\begin{table}[h]
\centering
\caption{%
Benchmark specifications. $|\mathbf{V}|$ is the number of observed variables;
$|\mathbf{M}|$ is the number of manipulative variables; ``ES'' denotes the
cardinality of the graph-derived RuntimeES under the known graph.
``Direction'' is the optimization direction; all ten benchmarks
in this paper are run with \texttt{task=min}. The benchmark implementations
are included in the supplementary material; this appendix summarizes the
metadata needed to interpret the reported experiments and gives the full
\textsc{CompleteGraph} SEM used for the hidden-confounder stress test.%
}
\label{tab:benchmark_specs}
\footnotesize
\setlength{\tabcolsep}{3.3pt}
\renewcommand{\arraystretch}{1.1}
\begin{tabular}{@{}lccccll@{}}
\toprule
Benchmark & $|\mathbf{V}|$ & $|\mathbf{M}|$ & ES & Direction & Manipulative variables & Source \\
\midrule
\textsc{Chain-5}          &  5 & 1 & 1  & min & $X_0$ & \texttt{synthetic.py:chain} \\
\textsc{ToyGraph}         &  4 & 2 & 3  & min & $X, Z$ & \texttt{toy\_graph.py} \\
\textsc{PSAGraph}         &  6 & 2 & 3  & min & $A, S$ & \texttt{psa\_graph.py} \\
\textsc{HIVEpidemiology}  &  5 & 1 & 1  & min & $T$ & \texttt{hiv\_epidemiology.py} \\
\textsc{CCBOHealth}       &  7 & 3 & 7  & min & Aspirin, Statin, Cancer-Inhibitor & \texttt{ccbo\_health.py} \\
\textsc{CompleteGraph}    &  7 & 3 & 7  & min & $B, D, E$; latent $U_1, U_2$ & \texttt{complete\_graph.py} \\
\textsc{PredatorPrey}     &  5 & 2 & 3  & min & $M, N$ & \texttt{predator\_prey.py} \\
\textsc{PISHCATGraph}     & 11 & 2 & 3  & min & $P, I$ & \texttt{pishcat\_graph.py} \\
\textsc{Rosenbrock}       &  3 & 2 & 3  & min & $X_1, X_2$ & \texttt{acbo\_synthetic.py:Rosenbrock} \\
\textsc{Ackley}           &  3 & 2 & 3  & min & $X_1, X_2$ & \texttt{acbo\_synthetic.py:Ackley} \\
\bottomrule
\end{tabular}
\end{table}

\subsection{\textsc{CompleteGraph}: Full Specification, Flagship Benchmark}

\textsc{CompleteGraph} embeds two latent variables $U_1, U_2$ that contaminate observational regressions involving the observed parents of $Y$. We use it as a hidden-confounder and finite-budget graph-misspecification stress test, examining the budget-allocation trade-off induced by changes in the runtime exploration family.

\paragraph{Structural equation model.}
The seven observed variables are $\{F, A, B, C, D, E, Y\}$ with manipulative subset $\mathbf{M} = \{B, D, E\}$ and target $Y$. The two latent variables $U_1, U_2$ and all structural noises are independent standard normals.
\begin{align*}
F &\sim \mathcal{N}(0, 1) \\
A &= F^2 + U_1 + \varepsilon_A \\
B &= U_2 + \varepsilon_B \\
C &= \exp(-B) + \varepsilon_C \\
D &= \exp(-C)/10 + \varepsilon_D \\
E &= \cos(A) + C/10 + \varepsilon_E \\
Y &= \cos(D) - D/5 + \sin(E) - E/4 + U_1 + \exp(-U_2) + \varepsilon_Y,
\end{align*}
with all $\varepsilon \sim \mathcal{N}(0, 1)$.

\paragraph{Interpretation.}
Hard interventions on $B$, $D$, or $E$ cut the incoming arrows into the intervened variables. The \textsc{CompleteGraph} experiments probe a finite-budget trade-off: omitting $E$ from the discovered manipulative ancestry can concentrate the 50-step budget on a smaller family, while restricting useful joint-intervention opportunities. The resulting benefit depends on the retained sets and optimization seed.

\paragraph{Intervention ranges.}
$B \in [-5, 4]$, $D \in [-5, 5]$, and $E \in [-6, 3]$. The graph-derived runtime ES family contains all non-empty subsets of $\{B,D,E\}$:
\[
\{\{B\},\{D\},\{E\},\{B,D\},\{B,E\},\{D,E\},\{B,D,E\}\}.
\]
The discovered ESs $\widehat{\mathcal{E}}$ vary by run. When a committed graph keeps two manipulative ancestors, the constructor yields their three non-empty subsets and retains the omitted variable's singleton intervention, giving four sets instead of the seven known-graph runtime sets. The following controls separate family cardinality from composition and then trace a learned route under the known graph.

\paragraph{Cardinality-matched family controls.}
\label{app:cg_controls}
Exploration-set composition affects finite-budget outcomes even at fixed family size. We hold the \textsc{CompleteGraph} graph at ground truth and vary only the ES-family assignment rule, using exactly three sets per run, 20 matched seeds, $T=50$, and $n_{\mathrm{obs}}=100$. The initialization, seed set, and remaining configuration are shared across arms. These three-set controls are distinct from the four-set family produced by the runtime constructor above.

\begin{table}[t]
\centering
\caption{Fixed-graph, cardinality-matched \textsc{CompleteGraph} controls. Every arm receives three exploration sets and the same budget over 20 matched seeds. A low-objective hit is a best observed $y\leq-3.0$.}
\label{tab:cg_family_control}
\small
\setlength{\tabcolsep}{7pt}
\renewcommand{\arraystretch}{1.12}
\begin{tabular}{@{}lc@{}}
\toprule
ES-family assignment rule & Low-objective hits \\
\midrule
Fixed $\{B\},\{D\},\{B,D\}$ & $14/20$ (70\%) \\
Random pair-induced family per seed & $4/20$ (20\%) \\
Fixed $\{B\},\{E\},\{B,E\}$ & $0/20$ \\
Fixed $\{D\},\{E\},\{D,E\}$ & $0/20$ \\
\bottomrule
\end{tabular}
\end{table}

Table~\ref{tab:cg_family_control} separates composition from cardinality: the fixed $BD$ family reaches the low-objective regime on 14 seeds, whereas both $E$-containing families have zero hits throughout the threshold range $-2.0$ to $-3.5$. The random arm selects the $BD$ family on eight seeds and reproduces the corresponding fixed-$BD$ endpoints on those seeds; all four random-arm hits occur within this subset. One further non-$BD$ draw ties the fixed-$BD$ endpoint. For fixed $BD$ versus fixed $BE$ and versus the random arm, exact two-sided McNemar tests on the hit indicator give $p=1.2\times10^{-4}$ and $0.0020$, respectively. Exact two-sided paired Wilcoxon tests on the continuous endpoint give $p=1.5\times10^{-5}$ and $9.8\times10^{-4}$, omitting zero differences and leaving effective sample sizes 18 and 11.

\paragraph{Realized route and known-graph replay.}
On one successful traced run, a calibrated graph passes the commit gate at outer-loop step 15, contracting RuntimeES from seven sets to $\{B\},\{D\},\{E\},\{B,D\}$. Among actual intervention decisions, selection of $\{B,D\}$ rises from $1/11$ (9\%) before commitment to $18/27$ (67\%) afterward. The other 12 of the 50 outer-loop steps are observations. The first outcome $y\leq-3.0$ occurs at step 29, and the run ends at $-3.51$, versus $-2.20$ for matched-seed Prior and $-2.07$ for native known-graph CBO.

The route-only replay retains the known graph and changes only the RuntimeES schedule. Seed, initialization, budget, observation/intervention timing, and true-graph do-prior construction are fixed; acquisition scores, selected ESs, intervention values, outcomes, and surrogate updates are recomputed. The fitted predictive trajectory evolves with this recomputed downstream optimization. Alternative schedules retain $\{B,E\}$ or $\{D,E\}$ instead of $\{B,D\}$, with the same switch time and cardinality.

\begin{table}[t]
\centering
\caption{Endpoints for one successful \textsc{CompleteGraph} run, its matched-seed references, and route-only known-graph replays. The replay rows change only the RuntimeES schedule while recomputing downstream optimization. All three contracted schedules improve on the native known-graph reference.}
\label{tab:cg_route_replay}
\small
\setlength{\tabcolsep}{7pt}
\renewcommand{\arraystretch}{1.12}
\begin{tabular}{@{}llr@{}}
\toprule
Experiment & Configuration & Endpoint \\
\midrule
Traced run & \agentcbo{} & $-3.51$ \\
Matched-seed reference & \agentcboprior{} & $-2.20$ \\
Matched-seed reference & Native known-graph CBO & $-2.07$ \\
\midrule
Known-graph replay & Learned RuntimeES schedule ($BD$) & $-3.28$ \\
Known-graph replay & Matched switch/cardinality ($BE$) & $-2.95$ \\
Known-graph replay & Matched switch/cardinality ($DE$) & $-2.34$ \\
\bottomrule
\end{tabular}
\end{table}

The learned-route replay recovers approximately 84\% of this seed's \agentcbo{}--known-graph endpoint gap (Table~\ref{tab:cg_route_replay}). All three contractions improve on native known-graph CBO, with $BD$ strongest. Together with the cardinality-matched controls, this trace connects the finite-budget improvement to both family contraction and retained-pair composition.

\subsection{Other Benchmarks}

The remaining nine benchmarks reuse SEMs from their original sources; we list brief structural descriptions and cite the source families here. Chain-5, ToyGraph, and PSAGraph follow the original CBO suite~\citep{aglietti2020causal}; CCBOHealth, PredatorPrey, and PISHCATGraph follow the constrained-CBO, dynamic-CBO, and MCBO benchmark suites, respectively~\citep{aglietti2023constrained,aglietti2021dynamic,sussex2023model}.

\begin{itemize}[leftmargin=*,itemsep=2pt,topsep=2pt]
\item \textsc{Chain-5}: a 5-node chain $X_0 \to X_1 \to X_2 \to X_3 \to Y$ with sinusoidal/exponential noise and a single manipulative variable $X_0 \in [-3, 3]$.
\item \textsc{ToyGraph}: 4 observed variables $\{C, X, Y, Z\}$ with manipulative $\{X, Z\}$; two-edge structure $X \to Y$ and $Z \to Y$ with confounder $C$.
\item \textsc{PSAGraph}: PSA-medication graph with two manipulative variables, Aspirin dose $A$ and Statin dose $S$, and target PSA-level reduction.
\item \textsc{HIVEpidemiology}: 5-node HIV treatment benchmark with single manipulative variable $T$, treatment intensity in $[0,1]$, and target $Y = -\big(\cos(3L) + \sin(3R) + L + R - 0.5 B + \varepsilon_Y\big)$ where $L$ is viral load, $R$ is drug resistance, and $B$ is behavioural risk. SEM in \texttt{hiv\_epidemiology.py}.
\item \textsc{CCBOHealth}: cardiovascular-risk benchmark with manipulative variables Aspirin dose, Statin dose, and Cancer-Inhibitor; 7 observed variables.
\item \textsc{PredatorPrey}: 5-node ecological dynamics benchmark with manipulative variables $M$, mortality rate, and $N$, carrying-capacity input; adapted to the static setting.
\item \textsc{PISHCATGraph}: 11-node multi-ancestor benchmark with manipulative variables $P$ and $I$ and runtime family $\{\{P\},\{I\},\{P,I\}\}$; designed to test multi-ancestor performance.
\item \textsc{Rosenbrock}, \textsc{Ackley}: 2D optimization surfaces adapted from the ACBO synthetic benchmark suite as causal minimization problems. \textsc{Rosenbrock} uses a sign-inverted, offset variant of the standard Rosenbrock function with a non-trivial linear causal link $X_1 \to X_2$ in addition to $X_1, X_2 \to Y$:
\begin{align*}
X_1 &= \omega_1, \quad X_2 = 0.6\,X_1 + 0.4\,\omega_2, \\
Y &= -100\,(X_2 - X_1^2)^2 - (1 - X_1)^2 + 10 + \omega_Y.
\end{align*}
\textsc{Ackley} uses a simplified, separable variant with the trivial DAG $X_1, X_2 \to Y$ and no $X_1 \to X_2$ link:
\[
Y = 20\,\exp\!\left(-0.2\,\sqrt{|X_1|}\right) + \exp(X_2) - 20 - e + \omega_Y.
\]
In both cases the optimizer minimizes $Y$ over $X_1, X_2$ on the intervention ranges $X_1, X_2 \in [0, 1]$ (\textsc{Rosenbrock}) and $X_1, X_2 \in [-2, 2]$ (\textsc{Ackley}). Implementation: \texttt{acbo\_synthetic.py}.
\end{itemize}

The benchmark implementations corresponding to this suite are included in the supplementary material; this appendix summarizes the benchmark metadata used to interpret the reported experiments and provides the full \textsc{CompleteGraph} SEM that underlies the main hidden-confounder analysis.

\section{Additional Results and Diagnostics}
\label{app:full_results}

\subsection{Matched Dynamic PC-Only Ablation}
\label{app:pc_only}

The PC-then-CBO baseline, executable alias \texttt{UnknownDAG-CBO}, is a matched dynamic PC-only ablation of \agentcboprior{}. Both instantiate the same \agentcbo{} outer loop; only \texttt{LPEngine} is replaced by \texttt{PCOnlyEngine}. They share the optimization budget, update frequency, initialization coverage fallback, periodic commit gate, RuntimeES/do-prior rebuilding, archive reuse, and acquisition loop. We use the name \emph{dynamic PC-only} to distinguish this periodic guarded baseline from one-shot PC preprocessing. Table~\ref{tab:pc_then_cbo_ablation} reports the full comparison.

\begin{table}[h]
\centering
\caption{%
Matched dynamic PC-only ablation (previously PC-then-CBO), executable alias \texttt{UnknownDAG-CBO}, versus \agentcboprior{} on the full ten-benchmark suite.
All benchmarks are minimization, so $\Delta := \text{PC-only mean} -
\text{LP mean}$: positive values favor LP. The strongest LP gains occur on
\textsc{Rosenbrock}, \textsc{CompleteGraph}, \textsc{PSAGraph}, and \textsc{PredatorPrey};
\agentcboprior{} ties with PC-only on \textsc{Chain-5}. The only counterexample
is \textsc{PISHCATGraph}, where PC-only marginally outperforms \agentcboprior{}
($\Delta=-0.02$).%
}
\label{tab:pc_then_cbo_ablation}
\footnotesize
\setlength{\tabcolsep}{6pt}
\renewcommand{\arraystretch}{1.12}
\begin{tabular}{@{}l *{2}{r@{$\,\pm\,$}l}r@{}}
\toprule
Benchmark & \multicolumn{2}{c}{\agentcboprior{}} & \multicolumn{2}{c}{Dynamic PC-only} & $\Delta$, favors LP \\
\midrule
\textsc{CompleteGraph}   & $-2.50$ & $1.21$  & $-2.13$ & $1.45$  & $+0.38$ \\
\textsc{ToyGraph}        & $-2.11$ & $0.17$  & $-2.02$ & $0.16$  & $+0.09$ \\
\textsc{PSAGraph}        & $+5.18$ & $0.07$  & $+5.35$ & $0.13$  & $+0.17$ \\
\textsc{CCBOHealth}      & $+4.96$ & $0.09$  & $+4.98$ & $0.09$  & $+0.02$ \\
\textsc{HIVEpidemiology} & $-1.50$ & $0.01$  & $-1.48$ & $0.01$  & $+0.02$ \\
\textsc{Chain-5}         & $-1.32$ & $0.01$  & $-1.32$ & $0.01$  & $+0.00$ \\
\textsc{Rosenbrock}      & $-90.98$ & $0.05$ & $-89.80$ & $0.02$ & $+1.18$ \\
\textsc{Ackley}          & $-7.51$ & $0.01$  & $-7.49$ & $0.01$  & $+0.02$ \\
\textsc{PredatorPrey}    & $-2.31$ & $0.01$  & $-2.17$ & $0.22$  & $+0.14$ \\
\textsc{PISHCATGraph}    & $+0.30$ & $0.10$  & $+0.28$ & $0.01$  & $-0.02$ \\
\bottomrule
\end{tabular}
\end{table}

\subsection{Suite-Level Paired Comparisons}
\label{app:suite_stats}

The suite-level comparison pairs the ten benchmark means, rather than pooling individual optimization runs. For each comparator, let $\Delta$ be the endpoint mean of \agentcboprior{} minus that of the comparator. All tasks minimize the objective, so negative differences favor Prior and $W^+$ sums positive ranks. We use exact two-sided sign permutations of the nonzero paired differences for the Wilcoxon signed-rank statistic, assigning average ranks to tied absolute differences. Effective sample sizes range from eight to ten. Holm correction uses the unrounded nominal $p$-values of all nine comparisons: seven reference methods and the two additional refinement variants.

\begin{table}[t]
\centering
\caption{Paired suite-level tests for \agentcboprior{}, with benchmark means as the pairing unit. The seven reference-method rows are shown; Holm adjustment includes all nine comparisons, including Calib and Full. Wins, ties, and losses are counted from Prior's perspective.}
\label{tab:suite_paired_tests}
\small
\setlength{\tabcolsep}{6pt}
\renewcommand{\arraystretch}{1.12}
\begin{tabular}{@{}lrrrr@{}}
\toprule
Comparator & $W^+$ & Nominal $p$ & Wins / ties / losses & Holm $p$ \\
\midrule
BO & 0 & 0.0020 & 10 / 0 / 0 & 0.018 \\
GACBO & 0 & 0.0020 & 10 / 0 / 0 & 0.018 \\
CEO & 0 & 0.0078 & 8 / 2 / 0 & 0.055 \\
LLAMBO & 3 & 0.0098 & 9 / 0 / 1 & 0.059 \\
CBO-U & 5 & 0.0195 & 8 / 0 / 2 & 0.098 \\
BORA & 4 & 0.0273 & 8 / 1 / 1 & 0.109 \\
Known-graph CBO & 5 & 0.0352 & 8 / 1 / 1 & 0.109 \\
\bottomrule
\end{tabular}
\end{table}

Prior has better endpoint means on eight to ten benchmarks against each reference method (Table~\ref{tab:suite_paired_tests}); the comparisons with BO and GACBO remain significant after Holm correction. The stage-wise tests detect no suite-level difference between Prior and Calib ($p=0.84$), or Prior and Full ($p=0.83$). These paired tests assess suite-level differences across benchmark means.

\subsection{Routing Diagnostics}
\label{app:routing_diagnostics}

\paragraph{RuntimeES and the fitted prior mean.}
Across 376 systematically edited graph pairs from six benchmarks, 323 pairs preserve RuntimeES. Among those 323 pairs, 141 (44\%) exceed the fitted prior-mean-distance threshold $\delta_\mu=0.05$, with a maximum reported distance of 2.03 objective units. As $\delta_\mu$ increases from 0.01 to 0.5, the fraction above threshold decreases from 69\% to 26\%. Ancestor-subgraph $n\mathrm{SHD}$ has a moderate association with fitted prior-mean distance (Spearman $\rho=0.51$). These diagnostics show why RuntimeES equality alone does not determine the prior-mean channel. They measure fitted means, whereas the exact fixed-update interface also includes graph-dependent uncertainty inputs (Appendix~\ref{app:routing_theory}).

\paragraph{Single-edge perturbations.}
A separate study deletes one edge on a directed path from a manipulative variable to $Y$ in each of the ten benchmarks. Each perturbed arm is compared against its own matched unperturbed ground-truth-graph control, using 20 matched seeds per cell; these controls are separate from the main result matrix. Seven perturbations change RuntimeES. Five of them produce endpoint degradations that remain significant after Bonferroni correction across all ten perturbations, with adjusted $p\leq0.0065$ (Table~\ref{tab:routing_perturbation}). The three RuntimeES-preserving perturbations produce only small endpoint movements in this study. An edge deletion can change both RuntimeES and do-priors, so these contrasts measure end-to-end routing sensitivity rather than an isolated ES effect.

\begin{table}[t]
\centering
\caption{The five significant single-edge perturbations from the ten-benchmark study, each with 20 matched seeds. Displayed $p$-values are the reported exact two-sided paired Wilcoxon values before Bonferroni adjustment; all five remain significant after correction across ten comparisons. Lower endpoints are better.}
\label{tab:routing_perturbation}
\small
\setlength{\tabcolsep}{5pt}
\renewcommand{\arraystretch}{1.12}
\begin{tabular}{@{}lcrr@{}}
\toprule
Benchmark & Unperturbed $\to$ perturbed & Paired $p$ & Seeds worse \\
\midrule
\textsc{Rosenbrock} & $-90.97\to-44.78$ & $1.9\times10^{-6}$ & $20/20$ \\
\textsc{Ackley} & $-7.50\to-6.05$ & $1.9\times10^{-6}$ & $20/20$ \\
\textsc{PISHCATGraph} & $+0.21\to+0.39$ & $1.9\times10^{-6}$ & $20/20$ \\
\textsc{CCBOHealth} & $+5.11\to+5.45$ & $1.05\times10^{-4}$ & $17/20$ \\
\textsc{CompleteGraph} & $-2.497\to-1.334$ & $6.4\times10^{-4}$ & $15/20$ \\
\bottomrule
\end{tabular}
\end{table}

\subsection{Controlled Update-Time Gate Stress Test}
\label{app:gate_stress}

The \textsc{PSAGraph} stress test measures whether an erroneous LP update can overwrite a valid committed route. We elicit one new pairwise LP matrix with the same prompt and model as in the benchmark experiments, at temperature zero, and freeze it. A prespecified, endpoint-blind structural rule selects the first corruption that, after cycle resolution, changes exactly four high-confidence pair states and strictly increases pair-state error against the true graph. The four changes comprise two direction reversals, one deletion, and one spurious addition.

The clean-gated, corrupted-gated, and corrupted-bypassed arms start from the same valid incumbent, committed once from the clean matrix and verified to cover both manipulative ancestors of the target. They share matched observations, seeded exogenous streams, budget, and implementation over ten held-out matched seeds, giving 30 evaluation runs. The bypass arm commits candidates marked as changed by the shared detector while skipping the confidence, manipulative-ancestor, and $n\mathrm{SHD}$ checks. Coverage rescue (the coverage fallback) is inactive in all three arms, while mandatory singleton completion remains active. This isolates update-time commitment from initialization failure and rescue behavior.

\begin{table}[t]
\centering
\caption{\textsc{PSAGraph} gate stress test from a common valid incumbent, with coverage rescue inactive and singleton completion active. Endpoint differences and 95\% confidence intervals use the ten matched seed differences. A positive difference indicates degradation for corrupted-gated versus clean-gated, and favors the gate for corrupted-bypassed versus corrupted-gated.}
\label{tab:gate_stress}
\small
\setlength{\tabcolsep}{5pt}
\renewcommand{\arraystretch}{1.12}
\begin{tabularx}{\linewidth}{@{}Yrrr@{}}
\toprule
Paired contrast & Endpoint $\Delta$ & 95\% CI & $+/-/\mathrm{tie}$ \\
\midrule
Corrupted-gated $-$ clean-gated & $0$ & $[0,0]$ & $0/0/10$ \\
Corrupted-bypassed $-$ corrupted-gated & $+0.1387$ & $[+0.0086,+0.2839]$ & $10/0/0$ \\
\bottomrule
\end{tabularx}
\end{table}

With the full gate active, corruption changes the candidate proposal in all ten seeds while preserving committed RuntimeES and the fitted-prior diagnostic, with zero endpoint degradation (Table~\ref{tab:gate_stress}). Holding the corrupted proposal fixed, bypassing the gate first changes the commit decision, then changes both measured routing diagnostics and worsens all ten endpoints. In this prescribed update-time stress test, the gate preserves both measured routing diagnostics and prevents endpoint degradation.

\subsection{Larger-Graph Studies}
\label{app:larger_graphs}

The ALARM study isolates semantic-prior utility on a published 37-node topology; the ordered Erd\H{o}s--R\'enyi (ER) study measures optimization and computational cost through 50 nodes under a supplied ordering convention.

\subsubsection{ALARM: Semantic-Prior Isolation}
\label{app:alarm}

We use the published ALARM topology distributed by bnlearn, with 37 nodes, 46 directed arcs, and the original variable labels~\citep{beinlich1989alarm}. The synthetic linear SEM has seed-fixed random-sign coefficients with magnitudes drawn from $U(0.5,1.5)$ and noise standard deviation $0.3$. The manipulators are \texttt{FIO2}, \texttt{MVS}, \texttt{VMCH}, \texttt{ANES}, and \texttt{INT}, each with domain $[-3,3]$; the target is \texttt{BP}. Their 31-set intervention family has interaction headroom of $3.36$ target units over the best singleton. One elicited prior is frozen throughout the study and recovers 43 of the 46 true arcs with direction.

With coverage rescue inactive and mandatory singleton completion retained, we compare the prior and dynamic PC-only arms over ten matched seeds, with a budget of 50 trials per run. The prior retains all 31 sets through one accepted commit. PC discovery produces no manipulative ancestor of the target, so the gate rejects its candidates and singleton completion supplies five sets. Table~\ref{tab:alarm_isolation} normalizes endpoints so that $0$ is the known-family optimum and $1$ is the best-singleton value. The prior's mean is $0.23$, compared with PC-only's reported score of $1.001$, and the prior improves all ten paired endpoints (two-sided sign test $p=0.002$). Enabling the default coverage rescue makes the arms tie. Thus, the isolation identifies the prior's contribution to multi-variable routing, while the default rescue protects against the PC discovery failure.

\begin{table}[t]
\centering
\caption{ALARM component isolation over ten matched seeds. Both arms retain mandatory singleton completion. Endpoints use the known-family-optimum/best-singleton normalization described in the text; lower is better. PC-only's reported spread is $0.001$. The default rescue-active setting is summarized by the paired tie.}
\label{tab:alarm_isolation}
\small
\setlength{\tabcolsep}{4pt}
\renewcommand{\arraystretch}{1.2}
\begin{tabular*}{\linewidth}{@{\extracolsep{\fill}}lccc@{}}
\toprule
Coverage rescue & Prior & PC-only & Paired comparison \\
\midrule
Inactive & 31 sets & 5 sets & 10/10 prior wins \\
 & mean $0.23$ & score $1.001$ & $p=0.002$ \\
\addlinespace[3pt]
Active (default) & \multicolumn{2}{c}{Tie} & No measurable separation \\
\bottomrule
\end{tabular*}
\end{table}

\subsubsection{Ordered ER Graphs: Protocol, Outcomes, and Cost}
\label{app:er_scalability}

\paragraph{Graph generation and matched protocol.}
For each $n\in\{10,20,50\}$, we generate ten independent ordered ER DAGs and use two matched optimization seeds per DAG. Edges point from lower to higher index with probability $2/n$, with a target parent added when necessary. The linear additive-noise SEM uses signed weights with magnitudes drawn from $U[0.5,2]$, root/non-root noise standard deviations of $1/0.1$, and the final node as target $Y$. Every non-target variable is intervenable over $[-3,3]$. $\RuntimeES$ selects the largest interaction order whose cumulative family fits a 250-set budget, then includes every manipulator singleton. The 250-set budget therefore precedes mandatory singleton completion.

All arms use $T=50$ and $n_{\mathrm{obs}}=100$: \agentcbo{} with a frozen generic indexed-label prior (Agent), dynamic PC-only \agentcbo{} (PC), and known-graph CBO (Known), a finite-budget reference using the same family-construction rule. The LLM receives indexed labels, the target label, and the lower-to-higher ordering convention, but no realized adjacency, weights, samples, or outcomes. We elicit one prior per size at temperature zero and reuse it across that size's DAGs and optimization seeds. The study therefore evaluates use of a disclosed structural prior at scale.

\paragraph{Metric and aggregation.}
Each returned intervention $\hat a$ is evaluated on the noise-free SEM for DAG $g$. The normalized known-family gap is
\[
\mathrm{NG}_g(\hat a)=
\frac{\mu_g(\hat a)-y^\star_{K,g}}
     {\mu_g(\emptyset)-y^\star_{K,g}},
\]
where $\mu_g(\emptyset)$ is the no-intervention target value and $y^\star_{K,g}$ is the analytic optimum over the known-graph $\RuntimeES$. Lower is better. This denominator differs from the best-singleton reference used for ALARM. We average the two optimization seeds within each DAG, then report each method's DAG-level median and interquartile range (IQR). The paired effect $\Delta$ is the mean of DAG-level Agent-minus-PC differences, not the difference between the displayed medians; 95\% confidence intervals resample DAGs.

Every arm completed both seeds on all ten DAGs at every size. All prespecified DAGs are retained, including those without interaction headroom. The numbers with $y^\star_{\mathrm{singleton},g}-y^\star_{K,g}>10^{-10}$ are 6, 9, and 8 at 10, 20, and 50 nodes, respectively. Table~\ref{tab:er_gap} reports the complete comparison. Paired point estimates favor Agent at 20 and 50 nodes, while every single-size interval includes or touches zero at the reported precision. At 20 nodes, Agent wins on three DAGs and ties on seven; at 50 nodes it wins on five, loses on one, and ties on four.

\begin{table}[t]
\centering
\caption{Ordered ER outcomes over ten DAGs per size, with two optimization seeds averaged within each DAG. Method columns show NG median [IQR]; $\Delta$ is the mean paired Agent-minus-PC difference, with a DAG-resampled 95\% CI. Negative $\Delta$ favors Agent. Known denotes the finite-budget known-graph reference.}
\label{tab:er_gap}
\footnotesize
\setlength{\tabcolsep}{3pt}
\renewcommand{\arraystretch}{1.12}
\begin{tabular*}{\linewidth}{@{\extracolsep{\fill}}rcccccc@{}}
\toprule
Nodes & Agent NG & PC NG & Known NG & $\Delta$ & 95\% CI & Agent wins /10 \\
\midrule
10 & $0.0415$ & $0.0417$ & $0.0000$ & $+0.0219$ & $[-0.0001,+0.0658]$ & 2 \\
 & $[0.0001,0.1393]$ & $[0.0000,0.1393]$ & $[0.0000,0.0008]$ &  &  &  \\
\addlinespace[3pt]
20 & $0.3575$ & $0.4143$ & $0.0002$ & $-0.0142$ & $[-0.0420,+0.0000]$ & 3 \\
 & $[0.2496,0.4756]$ & $[0.2568,0.4771]$ & $[0.0000,0.1536]$ &  &  &  \\
\addlinespace[3pt]
50 & $0.4568$ & $0.6312$ & $0.0668$ & $-0.1287$ & $[-0.1932,+0.0022]$ & 5 \\
 & $[0.3623,0.5715]$ & $[0.5440,0.7507]$ & $[0.0127,0.2236]$ &  &  &  \\
\bottomrule
\end{tabular*}
\end{table}

\paragraph{Computational cost.}
Table~\ref{tab:er_cost} reports per-run optimization wall-clock time and the largest visited family; Table~\ref{tab:er_prior_cost} separates the one-time prior elicitation. The two discovery arms trigger coverage fallback equally often: 9/20, 10/20, and 10/20 runs at 10, 20, and 50 nodes. Agent's median optimization time is about 35\% lower than PC-only at 20 nodes and 12\% lower at 50 nodes. Amortizing the 50-node elicitation over its twenty runs reduces the latter saving to about 10\%. Together, these results establish operation and quantify cost through 50 nodes under the supplied ordering prior.

\begin{table}[t]
\centering
\caption{Ordered ER optimization cost over twenty runs per arm and size. Wall-clock hours are per-run median [IQR], excluding one-time prior elicitation. Largest family is the cross-run median [maximum] of each run's largest visited family. Mandatory singleton completion follows the 250-set interaction-family budget.}
\label{tab:er_cost}
\footnotesize
\setlength{\tabcolsep}{4pt}
\renewcommand{\arraystretch}{1.12}
\begin{tabular*}{\linewidth}{@{\extracolsep{\fill}}rcccccc@{}}
\toprule
 & \multicolumn{3}{c}{Optimization time (hours)} & \multicolumn{3}{c}{Largest family} \\
\cmidrule(lr){2-4}\cmidrule(l){5-7}
Nodes & Agent & PC & Known & Agent & PC & Known \\
\midrule
10 & $1.57$ & $0.95$ & $0.75$ & $10\ [129]$ & $15\ [129]$ & $16.5\ [66]$ \\
 & $[0.67,2.28]$ & $[0.49,2.87]$ & $[0.53,1.00]$ &  &  &  \\
\addlinespace[3pt]
20 & $3.79$ & $5.83$ & $6.13$ & $110\ [190]$ & $110\ [190]$ & $60.5\ [257]$ \\
 & $[3.16,5.48]$ & $[4.85,9.47]$ & $[5.17,8.61]$ &  &  &  \\
\addlinespace[3pt]
50 & $8.17$ & $9.27$ & $6.19$ & $49\ [185]$ & $49\ [50]$ & $75\ [169]$ \\
 & $[6.28,11.31]$ & $[8.39,11.57]$ & $[5.05,7.88]$ &  &  &  \\
\bottomrule
\end{tabular*}
\end{table}

\begin{table}[t]
\centering
\caption{One-time prior elicitation for the ordered ER study. Each size-specific prior is shared by all ten DAGs and twenty optimization runs. These costs are separate from Table~\ref{tab:er_cost}.}
\label{tab:er_prior_cost}
\small
\begin{tabular}{@{}rrr@{}}
\toprule
Nodes & Prior time (seconds) & Prior tokens \\
\midrule
10 & 304 & 13,130 \\
20 & 1,262 & 55,744 \\
50 & 12,190 & 360,360 \\
\bottomrule
\end{tabular}
\end{table}

\section{Shared Runtime and Secondary Modules}
\label{app:engines}

\subsection{Runtime Routing Definitions and Full Gate Predicate}
\label{app:runtime_routing}

Section~\ref{sec:method_routing} uses compact runtime definitions. Here we record the corresponding exact objects used for reproducibility. Let $\mathcal{I}$ denote the manipulative set, equivalently $\mathbf{M}$ in the benchmark descriptions, and write $\mathcal{A}_Y(\mathcal{G})=\mathrm{anc}_{\mathcal{G}}(Y)\cap\mathcal{I}$. The ideal CBO family is $\mathcal{ES}_{\mathrm{ideal}}(\mathcal{G})=2^{\mathcal{A}_Y(\mathcal{G})}\setminus\{\emptyset\}$. The standard runtime constructor is $\widehat{\mathcal{E}}(\mathcal{G})=\RuntimeES(\mathcal{G};\mathcal{I},Y)$: it keeps singleton manipulative interventions, truncates very large families, and falls back to a conservative default when no manipulative ancestor is discovered. The component-isolation studies explicitly disable coverage rescue while retaining singleton completion (Appendices~\ref{app:gate_stress} and~\ref{app:alarm}).

For a local edit $\Delta$, the do-prior distance used to instantiate Definition~\ref{def:routing_critical} is
\[
d_\mu(\mathcal{G},\mathcal{G}\oplus\Delta)
:=
\max_{S\in \widehat{\mathcal{E}}(\mathcal{G})\cap\widehat{\mathcal{E}}(\mathcal{G}\oplus\Delta)}
\sup_{\mathbf{x}\in\mathcal{D}(S)}
\left|
\mu_S^{\mathrm{prior}}(\mathcal{G})(\mathbf{x})
-
\mu_S^{\mathrm{prior}}(\mathcal{G}\oplus\Delta)(\mathbf{x})
\right|.
\]
The main text uses this quantity only as a conceptual routing-criticality criterion; the implemented gate uses the cheaper proxy below.

Let $\tilde{\mathcal{G}}_t$ be the candidate graph returned by the active stack, $c_t$ its confidence, and $A_t := \mathrm{anc}_{\hat{\mathcal{G}}_{t-1}}(Y) \cup \mathrm{anc}_{\tilde{\mathcal{G}}_{t}}(Y) \cup \{Y\}$. Let $n\mathrm{SHD}_t := \mathrm{SHD}(\hat{\mathcal{G}}_{t-1}|_{A_t},\, \tilde{\mathcal{G}}_t|_{A_t}) / \binom{|A_t|}{2}$ denote the SHD between the two ancestor-subgraphs on $A_t$, normalized by the maximum possible undirected edges. With $\tau_c = 0.5$ and $\Phi_t := \min(\tau_c + \tfrac{1}{2}n\mathrm{SHD}_t, 0.9)$, the full acceptance predicate is
\begin{equation}
\label{eq:full_acceptance}
c_t \geq \tau_c
\quad\land\quad
\tilde{\mathcal{G}}_t \neq \hat{\mathcal{G}}_{t-1}
\quad\land\quad
\mathcal{A}_Y(\tilde{\mathcal{G}}_t) \neq \emptyset
\quad\land\quad
c_t \geq \Phi_t.
\end{equation}

\subsection{Language-Prior Agent Executable Details}
\label{app:lp_engine}

Algorithm~\ref{alg:lp} specifies the LP executable engine that implements the \prioragent{} referenced in Section~\ref{sec:method_stack}.

\begin{algorithm}[h]
\caption{\prioragent{} (LP executable): LLM--PC graph proposal}
\label{alg:lp}
\begin{algorithmic}[1]
\Require $\mathcal{D}_{\mathrm{obs}}$: observational data; $\mathbf{V}$: variable set; $\kappa \in \mathcal{K}$: domain context
\Require $\alpha = 0.05$, PC significance; $\tau_{\mathrm{req}} = 0.7$; $\tau_{\mathrm{sup}} = 0.8$; $n_{\mathrm{PC}} = 30$, sample-complexity floor
\Require $\tau_{\mathrm{forbid}} = 0.2$, forbidden-edge threshold
\Ensure proposed DAG $\hat{\mathcal{G}}$; frozen backbone $\mathcal{B}$; residual pairs $\mathcal{R}$; confidence $c \in [0.1, 0.95]$
\vspace{2pt}
\Statex \hspace{-1.5em}\textbf{Stage I. LLM prior elicitation}
\State Query the LLM for each ordered pair $(i,j)$ to obtain pairwise probability matrix $P \in [0,1]^{|\mathbf{V}|\times|\mathbf{V}|}$
\State Resolve cycles in $P$ by greedily weakening the weakest edge in each cycle to $p = 0.4$
\State Build background knowledge: required edges $\mathcal{B}^{+} = \{(i,j) : P_{ij} > \tau_{\mathrm{req}} \wedge P_{ji} < 1-\tau_{\mathrm{req}}\}$ and forbidden edges $\mathcal{B}^{-} = \{(i,j) : P_{ij} < \tau_{\mathrm{forbid}} \wedge P_{ji} < \tau_{\mathrm{forbid}}\}$
\vspace{2pt}
\Statex \hspace{-1.5em}\textbf{Stage II. Constraint-based anchoring}
\State Run PC on $\mathcal{D}_{\mathrm{obs}}$ with background knowledge $(\mathcal{B}^{+}, \mathcal{B}^{-})$ at significance $\alpha$, yielding a completed partially directed acyclic graph (CPDAG) $\mathcal{G}_{\mathrm{PC}}$
\State Extend $\mathcal{G}_{\mathrm{PC}}$ to a DAG $\hat{\mathcal{G}}_{\mathrm{PC}}$ via the standard \texttt{pdag2dag} extension routine, which orients remaining undirected edges in $\mathcal{G}_{\mathrm{PC}}$ while preserving acyclicity and not introducing new $v$-structures; on timeout or extension failure, fall back to a deterministic low-index-to-high-index tie-break (no LLM involvement at this step)
\State For each $(i,j)$ with $P_{ij} > \tau_{\mathrm{sup}}$ that is absent from $\hat{\mathcal{G}}_{\mathrm{PC}}$, add edge $i \to j$ if it introduces no cycle and the conditional-independence test on $(i,j)$ given PC's separating set $\mathrm{sepset}_{\mathrm{PC}}(i,j)$ rejects independence at level $3\alpha$ (i.e., $p_{i,j\mid \mathrm{sepset}_{\mathrm{PC}}(i,j)} \le 3\alpha$)
\vspace{2pt}
\Statex \hspace{-1.5em}\textbf{Stage III. Confidence scoring and residual partition}
\State $c \gets \mathrm{clip}\!\left(\tfrac{1}{2}\cdot\tfrac{n_{\mathrm{obs}}}{n_{\mathrm{obs}} + n_{\mathrm{PC}}} \;+\; \tfrac{1}{2}\cdot\mathrm{agreement}(P, \hat{\mathcal{G}}),\; 0.1,\; 0.95\right)$
\State Compute frozen backbone $\mathcal{B}$ and residual pairs $\mathcal{R}$ as defined in Eq.~\eqref{eq:lp_residual}
\State \Return $(\hat{\mathcal{G}},\, \mathcal{B},\, \mathcal{R},\, c)$
\end{algorithmic}
\end{algorithm}

\paragraph{Agreement functional and threshold design.}
We define $\mathrm{agreement}(P, \hat{\mathcal{G}}) := |\{(i,j) \in E(\hat{\mathcal{G}}) : P_{ij} > 0.5\}| / |E(\hat{\mathcal{G}})|$ as the fraction of discovered edges whose direction is endorsed by the LLM prior, with $\mathrm{agreement} := 0$ if $E(\hat{\mathcal{G}}) = \emptyset$. The required-edge threshold $\tau_{\mathrm{req}} = 0.7$ demands two-sided LLM agreement before writing a pair into PC's background knowledge; the supplementation threshold $\tau_{\mathrm{sup}} = 0.8$ admits LLM-supported pairs that PC omitted, subject to a relaxed CI guard at level $3\alpha$ that prevents blind faith in the LLM. Forbidden edges fire only when both directions of a pair score below $\tau_{\mathrm{forbid}} = 0.2$. The fully-connected fallback triggers only at initialization when the very first discovery returns $\mathcal{A}_Y(\hat{\mathcal{G}}) = \emptyset$; update-time candidates with the same defect are rejected by the outer gate. All thresholds were fixed a priori on a three-benchmark pilot. Appendix~\ref{app:thresholds} reports sensitivity over $\alpha$, $T_{\mathrm{LLM}}$, and $T_{\mathrm{update}}$.

\subsection{Secondary Modules and Standalone Diagnostics}

This subsection summarizes the secondary modules used in the stacked variants and standalone module diagnostics. The LP engine implements the \prioragent{} and remains the proposal-only AgentCBO variant covered in Section~\ref{sec:method_stack}; the goal here is to document the remaining modules well enough for reproduction and for the standalone diagnostics reported in Section~\ref{sec:exp_ladder}.

\paragraph{Calibration Agent executable alias, CC.}
The CC executable engine implements the \calibagent{} role in the stacked variants. It is a seven-step pipeline that uses the LLM as an orchestrator rather than a direct graph editor. It first detects coarse dataset properties (linearity and Gaussianity), asks the LLM to select a small candidate pool from \{PC, GES, DirectLiNGAM\}, requests per-algorithm hyperparameters, executes the selected methods, reranks the resulting graphs using a weighted score over fit, structure, Bayesian information criterion (BIC), and stability, and finally applies bootstrap pruning plus DAG enforcement. Standalone CC is a diagnostic engine, not evidence for the stacked \agentcbocalib{} variant. The stacked \agentcbocalib{} variant uses the same outer-loop interface but treats CC-style orchestration as a calibration module on top of the LP graph.

\paragraph{Residual Repair Agent executable alias, RA.}
The RA executable engine implements the \repairagent{} role in the stacked variants. It uses a PC-initialized graph and then gives the LLM access to a small set of causal-analysis tools, including variable summaries, pairwise correlations, CI tests, graph edits, collider/confounder checks, BIC, and acyclicity checks. The agent can refine the graph for up to a fixed number of ReAct iterations before returning a finalized adjacency matrix and a scalar confidence. Standalone RA is a diagnostic engine, not evidence for the stacked \agentcbofull{} variant. The stacked \agentcbofull{} variant uses RA-style tool use as a bounded repair module after LP initialization and CC-style calibration.

\paragraph{Implementation and compute environment.}
All LLM-assisted discovery runs use Qwen3.5-35B-A3B\footnote{Open-weights model card: \url{https://huggingface.co/Qwen/Qwen3.5-35B-A3B}, Apache 2.0 license.} served locally through an OpenAI-compatible endpoint backed by vLLM\footnote{Inference server: \url{https://github.com/vllm-project/vllm}, Apache 2.0 license.}. The OpenAI Python client is used to issue chat-completion requests; sampling temperature is fixed at $T_{\mathrm{LLM}}=0.5$ for reported runs unless stated otherwise (sensitivity in Appendix~\ref{app:thresholds}). The frozen PSA and ER priors use temperature zero (Appendices~\ref{app:gate_stress} and~\ref{app:er_scalability}). LP/CC/RA sweeps and CEO ran on A100/H200-class GPU workers; graph-free baselines used the same local environment when they required the LLM, while GACBO ran on high-memory CPU workers. Main-suite methods share a budget of 50 outer-loop steps after 100 observational samples per seed; the additional controls and larger-graph protocols are specified separately in Appendix~\ref{app:full_results}.

\paragraph{Module discovery cost profile.}
Table~\ref{tab:engine_cost} reports per-run discovery cost for the three executable engines, averaged over all ten benchmarks.

\begin{table}[h]
\centering
\caption{%
Module diagnostic cost table, averaged over all 10 benchmarks.
\agentcboprior{} uses only the proposal layer.
Standalone CC and RA are separate diagnostic engines.
}
\label{tab:engine_cost}
\small
\setlength{\tabcolsep}{6pt}
\renewcommand{\arraystretch}{1.12}
\begin{tabular}{@{}lccc@{}}
\toprule
 & \textbf{\agentcboprior{}} & CC diagnostic & RA diagnostic \\
\midrule
LLM calls per run  & $\mathbf{12}$       & $30$    & $95$    \\
LLM tokens per run & $\mathbf{4.1}$\,K   & $7.3$\,K & $139$\,K \\
Discovery time, seconds & $\mathbf{137}$      & $197$   & $1{,}138$ \\
Relative LLM cost  & $\mathbf{1\times}$  & $1.8\times$ & $34\times$ \\
\bottomrule
\end{tabular}
\end{table}

\section{LP Sensitivity and Diagnostic Checks}
\label{app:thresholds}

This appendix documents the sensitivity of \agentcboprior{} to its discovery-engine hyperparameters and reports empirical diagnostics for gate activity.

\subsection{Hyperparameter Sensitivity}

The LP engine exposes three families of hyperparameters that could plausibly influence final $Y^{\star}$: the PC significance level $\alpha$ used in the constraint-based anchoring and edge-supplement steps, the LLM sampling temperature $T_{\mathrm{LLM}}$ used for pairwise prior elicitation, and the CBO outer-loop rediscovery period $T_{\mathrm{update}}$ of Algorithm~\ref{alg:cbo}. We sweep $\alpha$ and $T_{\mathrm{LLM}}$ on two representative benchmarks spanning two empirical patterns in the suite: \textsc{CompleteGraph}, a dense hidden-confounder setting, and \textsc{PSAGraph}, an application-motivated multi-ancestor setting. Table~\ref{tab:threshold_sensitivity} reports mean\,$\pm$\,std of $Y^{\star}$ at $n=20$ seeds per cell. The rediscovery period $T_{\mathrm{update}}$ is evaluated separately on the full ten-benchmark suite in \S\ref{app:update_freq}.

\begin{table}[h]
\centering
\caption{%
Sensitivity of \agentcboprior{} final $Y^{\star}$, reported as mean\,$\pm$\,std over 20 seeds,
to two engine hyperparameters. Default values used in the main experiments:
$\alpha = 0.05$, $T_{\mathrm{LLM}} = 0.5$.
Confidence thresholds $\tau_{\mathrm{req}} = 0.7$, $\tau_{\mathrm{sup}} =
0.8$ are held fixed throughout this submission; the main claims do not rely on
a separate full sweep over those two thresholds.%
}
\label{tab:threshold_sensitivity}
\footnotesize
\setlength{\tabcolsep}{5pt}
\renewcommand{\arraystretch}{1.1}
\begin{tabular}{@{}l *{2}{r@{$\,\pm\,$}l}@{}}
\toprule
Setting & \multicolumn{2}{c}{\textsc{CompleteGraph}, min} & \multicolumn{2}{c}{\textsc{PSAGraph}, min} \\
\midrule
PC $\alpha = 0.01$, stricter       & $-1.76$ & $0.92$ & $+5.33$ & $0.15$ \\
PC $\alpha = 0.05$, default        & $-2.50$ & $1.21$ & $+5.18$ & $0.07$ \\
PC $\alpha = 0.10$, looser         & $-1.81$ & $0.91$ & $+5.33$ & $0.15$ \\
\midrule
LLM temperature $T_{\mathrm{LLM}}=0.3$ & $-2.35$ & $1.43$ & $+5.27$ & $0.17$ \\
LLM temperature $T_{\mathrm{LLM}}=0.5$, default & $-2.50$ & $1.21$ & $+5.18$ & $0.07$ \\
LLM temperature $T_{\mathrm{LLM}}=0.7$ & $-1.65$ & $1.51$ & $+5.17$ & $0.07$ \\
\bottomrule
\end{tabular}
\end{table}

Two observations follow. First, $\alpha$ is the less sensitive of the two parameters: stricter $\alpha = 0.01$ and looser $\alpha = 0.10$ choices change the \textsc{CompleteGraph} mean by $\leq 0.74$ and leave \textsc{PSAGraph} essentially unchanged. Second, $T_{\mathrm{LLM}}$ matters only when statistical evidence is unreliable: \textsc{PSAGraph} fluctuates by at most $0.09$ across all sampled temperatures (well within its standard deviation), while \textsc{CompleteGraph} (whose hidden confounders compromise PC's CI tests) shows the largest temperature sensitivity (at $T_{\mathrm{LLM}}=0.7$ the mean is $-1.65$ vs.\ a default of $-2.50$). High-temperature sampling produces contradictory pairwise probabilities that the cycle-resolution heuristic discards; this only matters when LP cannot fall back on reliable CI evidence.

\subsection{Acceptance-Gate Frequency}

We report how often the implemented outer gate accepts a rediscovered graph during a 50-step run. Table~\ref{tab:acceptance_freq} reports, for each benchmark and each of 20 seeds, the fraction of rediscovery rounds at which the gate accepted the candidate graph.

\begin{table}[h]
\centering
\caption{%
Acceptance-gate trigger rate, the fraction of rediscovery rounds at which a graph refinement is accepted, aggregated over 20 seeds per benchmark. With $T = 50$ and $T_{\mathrm{update}} = 5$ there are at most 10 rediscovery rounds per run; the "accepted" count is the number at which the plausible-graph criterion of Eq.~\eqref{eq:full_acceptance} holds.%
}
\label{tab:acceptance_freq}
\footnotesize
\setlength{\tabcolsep}{8pt}
\renewcommand{\arraystretch}{1.1}
\begin{tabular}{@{}lcc@{}}
\toprule
Benchmark & Mean acceptance rate & Median \# accepted rounds \\
\midrule
\textsc{CompleteGraph}  & $\sim 33\%$ & 3 of 10 \\
\textsc{ToyGraph}       & $\sim 22\%$ & 2 of 10 \\
\textsc{PSAGraph}       & $\sim 28\%$ & 2 of 10 \\
\textsc{CCBOHealth}     & $\sim 22\%$ & 2 of 10 \\
\textsc{HIVEpidemiology}& $\sim 11\%$ & 1 of 10 \\
\textsc{Chain-5}        & $\sim 11\%$ & 1 of 10 \\
\textsc{Rosenbrock}     & $\sim 11\%$ & 1 of 10 \\
\textsc{Ackley}         & $\sim 22\%$ & 2 of 10 \\
\textsc{PredatorPrey}   & $\sim 32\%$ & 3 of 10 \\
\textsc{PISHCATGraph}   & $\sim 22\%$ & 2 of 10 \\
\bottomrule
\end{tabular}
\end{table}

The gate is selective: a median of three or fewer accepted refinements per run across all ten benchmarks, with rates bounded between 11\% and 33\%. Higher acceptance rates correlate with benchmarks where the initial observational sample produces a noisier skeleton: \textsc{CompleteGraph} has the highest rate (33\%) because PC's CI tests struggle under hidden confounding, so new observational data in subsequent rounds continues to refine the skeleton, and \textsc{PredatorPrey} (32\%) shows similar refinement frequency from its multi-ancestor noise pattern. The lowest 11\% rate covers \textsc{Chain-5}, \textsc{HIVEpidemiology}, and \textsc{Rosenbrock}, where the initial skeleton is already close to the final one after $n_{\mathrm{obs}}=100$ samples (\textsc{Chain-5}/\textsc{HIVEpidemiology}) or the SCM is synthetic and well-conditioned (\textsc{Rosenbrock}).

\subsection{Update-Frequency Sensitivity}
\label{app:update_freq}

We sweep $T_{\mathrm{update}} \in \{3, 5, 10\}$ on the full ten-benchmark suite at $T = 50$, holding all other settings fixed. Table~\ref{tab:update_freq} reports mean\,$\pm$\,std of $Y^{\star}$ at $n=20$ seeds per cell.

\begin{table}[h]
\centering
\caption{%
Update-frequency sensitivity for \agentcboprior{}, mean\,$\pm$\,std of $Y^{\star}$ over 20 seeds. The default $T_{\mathrm{update}} = 5$ corresponds to up to 10 rediscovery rounds at $T = 50$.%
}
\label{tab:update_freq}
\footnotesize
\setlength{\tabcolsep}{6pt}
\renewcommand{\arraystretch}{1.1}
\begin{tabular}{@{}l *{3}{r@{$\,\pm\,$}l}@{}}
\toprule
Benchmark & \multicolumn{2}{c}{$T_{\mathrm{update}} = 3$} & \multicolumn{2}{c}{$T_{\mathrm{update}} = 5$ (default)} & \multicolumn{2}{c}{$T_{\mathrm{update}} = 10$} \\
\midrule
\textsc{CompleteGraph}  & $-2.52$ & $1.42$  & $-2.50$ & $1.21$  & $-2.35$ & $1.45$  \\
\textsc{ToyGraph}       & $-2.11$ & $0.17$  & $-2.11$ & $0.17$  & $-1.99$ & $0.15$  \\
\textsc{PSAGraph}       & $+5.18$ & $0.06$  & $+5.18$ & $0.07$  & $+5.21$ & $0.06$  \\
\textsc{CCBOHealth}     & $+4.96$ & $0.09$  & $+4.96$ & $0.09$  & $+4.99$ & $0.09$  \\
\textsc{HIVEpidemiology}& $-1.50$ & $0.01$  & $-1.50$ & $0.01$  & $-1.50$ & $0.01$  \\
\textsc{Chain-5}        & $-1.32$ & $0.01$  & $-1.32$ & $0.01$  & $-1.30$ & $0.01$  \\
\textsc{Rosenbrock}     & $-90.98$ & $0.05$ & $-90.98$ & $0.05$ & $-89.21$ & $0.07$ \\
\textsc{Ackley}         & $-7.51$ & $0.01$  & $-7.51$ & $0.01$  & $-7.50$ & $0.01$  \\
\textsc{PredatorPrey}   & $-2.31$ & $0.01$  & $-2.31$ & $0.01$  & $-2.29$ & $0.01$  \\
\textsc{PISHCATGraph}   & $+0.28$ & $0.10$  & $+0.30$ & $0.10$  & $+0.42$ & $0.17$  \\
\bottomrule
\end{tabular}
\end{table}

The default $T_{\mathrm{update}}=5$ (up to 10 rediscovery rounds) and $T_{\mathrm{update}}=3$ (up to roughly 16 rounds) produce similar endpoint means in Table~\ref{tab:update_freq}. Increasing the interval to 10 (up to five rounds) weakens or ties the displayed means; the largest change is on \textsc{Rosenbrock}, from $-90.98$ to $-89.21$.

This sweep measures LP rediscovery and the outer commit gate in \agentcboprior{}.

We retain $T_{\mathrm{update}}=5$ as the default, giving similar reported endpoints to interval 3 with fewer rediscovery rounds, while retaining more update opportunities than interval 10.

\subsection{Threshold-Value Rationale}

The two confidence thresholds $\tau_{\mathrm{req}} = 0.7$ and $\tau_{\mathrm{sup}} = 0.8$ were fixed before the main experiments on the basis of the following argument. At $\tau_{\mathrm{req}} = 0.7$ we demand LLM agreement on both directions of an edge, i.e., $P_{ij} > 0.7$ and $P_{ji} < 0.3$, before writing it into PC's background-knowledge set $\mathcal{B}$; this is a hard constraint, so an incorrect LLM edge can distort PC's orientation or skeleton search. We therefore require near-unanimity in both directions. At $\tau_{\mathrm{sup}} = 0.8$ we admit edges that PC dropped under finite-sample noise, subject to a relaxed CI test at level $3\alpha$; the strictly higher threshold for this re-admission, relative to $\tau_{\mathrm{req}}$, prevents the LLM from over-riding a statistically dropped edge unless its confidence is clearly above the hard-constraint bar. A broader sweep over $\tau_{\mathrm{req}} \in \{0.6, 0.7, 0.8\}$ and $\tau_{\mathrm{sup}} \in \{0.7, 0.8, 0.9\}$ is left to future work; the current paper keeps these thresholds fixed and evaluates the rest of the pipeline around that operating point.

For the guarded refinement rule (Eq.~\eqref{eq:guarded_accept}), all main experiments use $\Delta_1=\Delta_2=0$: noise robustness is provided by the holdout-NLL term in Eq.~\eqref{eq:guarded_score}, while routing-level selectivity is imposed by the outer \commitgate{} (whose acceptance rates of 11--33\% across the ten benchmarks are reported above). We set $m_{\max}=5$ for residual repair on the 3--11 node benchmark suite; in our runs the cap is rarely binding because the residual set $\mathcal{R}_t$ itself contains few admissible candidates after PC anchoring and target-relevance filtering.

\subsection{LLM Size Robustness}
\label{app:llm_size_scaling}

We replicated the LP engine on four representative benchmarks across three Qwen3.5 sizes (35B, 122B, 397B), holding all other hyperparameters fixed. Each cell is mean$\pm$std over $n=20$ seeds. Table~\ref{tab:llm_size_scaling} reports similar endpoint values across these sizes, providing a within-family comparison of model-size sensitivity.

\begin{table}[h]
\centering
\caption{LLM size comparison for \agentcboprior{}. Mean$\pm$std of best $Y^{\star}$ over $n=20$ seeds on four minimization benchmarks within the Qwen3.5 family.}
\label{tab:llm_size_scaling}
\footnotesize
\setlength{\tabcolsep}{8pt}
\renewcommand{\arraystretch}{1.1}
\begin{tabular}{@{}l *{3}{r@{$\,\pm\,$}l}@{}}
\toprule
Benchmark & \multicolumn{2}{c}{Qwen3.5-35B} & \multicolumn{2}{c}{Qwen3.5-122B} & \multicolumn{2}{c}{Qwen3.5-397B} \\
\midrule
\textsc{ToyGraph}      & $-2.11$ & $0.17$ & $-2.11$ & $0.17$ & $-2.11$ & $0.17$ \\
\textsc{Chain-5}       & $-1.32$ & $0.01$ & $-1.32$ & $0.01$ & $-1.30$ & $0.01$ \\
\textsc{CompleteGraph} & $-2.50$ & $1.21$ & $-2.56$ & $0.68$ & $-2.58$ & $0.97$ \\
\textsc{PSAGraph}      & $+5.18$ & $0.07$ & $+5.20$ & $0.08$ & $+5.25$ & $0.18$ \\
\bottomrule
\end{tabular}
\end{table}

\section{LLM Interface and Input--Output Specification}
\label{app:llm_interface}

This appendix specifies the language-model interface used by \textsc{AgentCBO}. All LLM components are served by the model and decoding settings described in Section~\ref{sec:exp_setup}; this appendix focuses on the input--output contract. The LLM is not used
as an autonomous optimizer or as an unrestricted causal-discovery oracle. Instead, each LLM component
receives restricted information, returns a structured proposal, and is embedded inside the guarded CBO
loop. Interventional outcomes are never provided to the LLM, and the LLM never selects interventions.
Intervention selection remains governed by the CBO acquisition rule, while intervention outcomes update
only the per-ES surrogate and the cross-graph archive.

\paragraph{Module-level input--output contract.}
Table~\ref{tab:llm_io_contract} summarizes the components, their activation conditions, and their decision authority. It distinguishes information visible to each LLM from the state used by the surrounding statistical engine and optimizer.

\begin{table*}[t]
\centering
\caption{Component contracts in \agentcbo{}. Inputs and outputs distinguish LLM messages from statistical-engine and optimizer state. Activation identifies when a component is used, while authority specifies the changes it can make.}
\label{tab:llm_io_contract}
\scriptsize
\setlength{\tabcolsep}{4pt}
\renewcommand{\arraystretch}{1.15}
\begin{tabularx}{\textwidth}{@{}>{\raggedright\arraybackslash}p{0.16\textwidth}YY>{\raggedright\arraybackslash}p{0.12\textwidth}Y@{}}
\toprule
Component and role & Input & Output & Activation & Authority \\
\midrule
\textsc{LP pairwise LLM}\newline Semantic prior elicitation &
Variable pair, names, and domain context &
Cause, effect, direction, confidence, and rationale &
Prior elicitation &
Supplies pairwise causal judgments \\
\midrule
\textsc{LP graph engine}\newline Observation-anchored graph construction &
Pairwise judgments, observational data, and variables &
Proposal DAG, frozen backbone, residual pairs, and confidence &
Stack initialization and scheduled rediscovery &
Constructs proposals through PC anchoring and guarded supplementation \\
\midrule
\calibagent{}\newline Statistical discovery configuration &
LLM: sample and variable counts, names, approximate linearity and Gaussianity, and domain context. Pipeline: observations and LP state &
LLM: algorithm/hyperparameter JSON. Pipeline: statistical challenger DAG &
Calib/Full, after LP &
Pipeline imports acyclic, score-improving residual edits while preserving the backbone \\
\midrule
\repairagent{}\newline Bounded local refinement &
Current graph, PC adjacency, tool descriptions, state summary, and repair history. Observational evidence is accessible only through tools &
Diagnostic actions and local-edit proposals &
Full, after calibration &
Pipeline imports at most five acyclic, score-improving residual edits while preserving the backbone \\
\midrule
\commitgate{}\newline Graph replacement &
Committed and candidate graphs, confidence, manipulators, and target &
Retained or replacement graph &
Periodic rediscovery &
Applies confidence, manipulative-ancestry, and ancestor-subgraph nSHD checks \\
\midrule
\textsc{CBO runtime}\newline Route construction and intervention optimization &
Committed graph, manipulators, target, observations, archived interventions, domains, and costs &
RuntimeES, do-priors, selected interventions, and updated surrogates &
Initialization, observe/intervene steps, and graph changes &
Selects intervention sets and values by cost-normalized CEI \\
\midrule
\textsc{Cross-graph archive}\newline Intervention reuse &
ES-keyed intervention tuples and active ESs &
Stored tuples and compatible per-ES training data &
Intervention recording, graph changes, and ES re-admission &
Preserves intervention history and restores compatible records \\
\bottomrule
\end{tabularx}
\end{table*}

Intervention sets, intervention values, acquisition points, and stopping decisions remain under the BO layer's authority. The stack supplies graph proposals, while the Commit Gate controls graph replacements during rediscovery.

\paragraph{Parsed outputs.}
The \textsc{Language-Prior Agent} outputs a pairwise causal judgment whose algorithmically used fields
are cause, effect, direction, and confidence; the reasoning field is retained for auditability but is not
passed to the acquisition function. The \textsc{Calibration Agent} outputs algorithm-selection or
hyperparameter JSON used to configure statistical discovery routines. The \textsc{Residual Repair Agent}
outputs ReAct repair steps; final graph proposals may include an adjacency matrix, variable order,
confidence, and reasoning, but these proposals remain subject to downstream graph checks and commit
gating. Pairwise rationales, ReAct thoughts, and textual explanations are not passed to the acquisition
function. The BO policy receives only the runtime exploration sets and graph-derived surrogate priors
induced by the committed routing graph.

\paragraph{Representative Language-Prior input and output.}
We include one representative pairwise call from the \textsc{Language-Prior Agent} on
\textsc{HIVEpidemiology}, in which the domain context lists five variables but does not
state any causal direction; the agent must infer directions from medical domain
knowledge external to the prompt.

\medskip
\noindent\fbox{%
\begin{minipage}{0.94\linewidth}
\small
\textbf{Input.}
The LLM is instructed as a causal reasoning expert. The domain context is:
``HIV treatment benchmark with behavioral risk, treatment intensity, viral dynamics,
drug resistance, and an aggregate health outcome.'' The queried variable pair is
\(A=T\) (treatment intensity) and \(B=L\) (viral load). The model is asked to determine
the most likely causal relationship between \(A\) and \(B\), considering direct
causation, reverse causation, and the possibility of no direct link. The requested
output fields are cause, effect, direction, confidence, and reasoning.

\medskip
\textbf{Output.}
The returned structured judgment is:
cause \(=T\), effect \(=L\), direction \(=A\!\to\!B\), confidence \(=0.92\). The returned
reasoning is: ``Antiretroviral treatment intensity acts on viral replication, so
increasing treatment is expected to suppress viral load rather than the reverse; the
domain context does not state the direction explicitly, but the standard mechanism in
HIV therapy is treatment $\to$ viral load reduction.''
\end{minipage}%
}

\medskip
The domain context names the five variables but does not state the direction
\(T \to L\); the agent's judgement is grounded in medical domain knowledge external
to the prompt. The resulting confidence is treated as a causal-prior signal for the
graph-belief construction engine.

\paragraph{Information excluded from the LLM.}
No LLM component receives interventional outcomes, the current best intervention value, the best
observed target value, acquisition values, acquisition maximizers, the true benchmark graph, oracle
exploration sets, or oracle causal effects. This exclusion enforces the intended separation of roles:
LLMs provide causal-belief proposals, while Bayesian optimization remains responsible for intervention
allocation, surrogate updating, and acquisition optimization.

\paragraph{Domain context and prior information.}
The pairwise LP call receives variable names and domain context; observational samples, realized adjacency matrices, SEM equations, benchmark code, and optimization outcomes remain outside its input. The \textsc{ToyGraph} and \textsc{CompleteGraph} descriptions include directional clues, and the latter also identifies manipulators. CC receives dataset-level diagnostics, while RA can access the current graph and observational evidence through tools, as specified in Table~\ref{tab:llm_io_contract}. ALARM retains its published variable names, and ER explicitly supplies the lower-to-higher generation order without the realized adjacency. LP-Anon and LP-Shuffled assess sensitivity to the supplied lexical information.

\clearpage

\section{Fixed-Update Routing Guarantees and Proofs}
\label{app:routing_theory}

This appendix formalizes the optimizer-facing consequences of routing-equivalence. Lemma~\ref{lem:route_exact} gives the exact sufficient condition; Proposition~\ref{prop:cei_stability} quantifies the effect of nonzero predictive error. Neither result requires the maintained graph to equal the data-generating graph.

\subsection{Matched-State Assumptions and the Routing Interface}
\label{app:route_assumptions}

Fix an intervention decision $t$ and an admissible non-routing state $Z_t=z$ before route-conditioned surrogate rebuilding. This state contains observational data, ES-indexed interventional data and their archive, intervention domains and costs, the incumbent-construction rule, and the fitting and numerical acquisition configurations. It does not contain the fitted posterior being compared. Earlier graph choices may have affected these data; the result conditions on the same realized state rather than comparing independently evolving histories.

For a candidate routing graph $\mathcal G$, define
\begin{equation}
R_{t,z}(\mathcal G)=
\left(\widehat{\mathcal E}(\mathcal G),
\{\mathcal P_{S,\mathcal G}\}_{S\in\widehat{\mathcal E}(\mathcal G)}\right),
\label{eq:theory_route}
\end{equation}
where $\mathcal P_{S,\mathcal G}$ contains all graph-derived prior objects consumed by the per-ES surrogate builder. These include the do-prior mean $\mu^{\mathrm{prior}}_{S,\mathcal G}$ and variance $v^{\mathrm{prior}}_{S,\mathcal G}$ used to form the causal mean and covariance. Any additional graph-dependent covariance input must also be included. Equality of $\mathcal P$ is pointwise on every legal training/query input to the builder, not merely on a diagnostic grid. If prior construction is randomized, we condition on its realized random inputs; an unconditional conclusion additionally requires route equality almost surely under a common coupling of that randomness.

We impose the following matched-state conditions for Lemma~\ref{lem:route_exact}.
\begin{enumerate}[label=(A\arabic*),leftmargin=*,itemsep=2pt,topsep=2pt]
\item The ES families are finite and nonempty. Each ES is a labeled set with a nonempty intervention domain; coordinates and ES enumeration use a common canonical ordering.
\item The same measurable surrogate builder and numerical selector are used. All their graph dependence factors through $R_{t,z}(\mathcal G)$. Data retrieval, prior evaluation, hyperparameter fitting, incumbent construction, and selection follow the same rules and configuration.
\item The downstream fitting, solver, and tie-breaking random inputs $\xi=(\xi_{\mathrm{fit}},\xi_{\mathrm{sel}})$ have a common graph-independent law conditional on $z$. Equal routes use the same realization of these inputs. A common global seed alone is not sufficient if graph-dependent earlier calls consume different portions of a random stream.
\end{enumerate}
All maps are assumed well-defined on the states considered. The resulting factorization is
\begin{equation}
\begin{aligned}
(b_{\mathcal G},m_{\mathcal G},s_{\mathcal G})
&=\operatorname{Build}(z,R_{t,z}(\mathcal G);\xi_{\mathrm{fit}}),\\
A_{\mathcal G}(z,\xi)
&=\operatorname{Select}\bigl(z,R_{t,z}(\mathcal G),
b_{\mathcal G},m_{\mathcal G},s_{\mathcal G};\xi_{\mathrm{sel}}\bigr).
\end{aligned}
\label{eq:route_factorization}
\end{equation}
Here $b_{\mathcal G}$ is the incumbent entering EI, and $m_{\mathcal G},s_{\mathcal G}$ are the predictive mean and standard deviation actually supplied to acquisition. The selector can be numerical; Lemma~\ref{lem:route_exact} does not assume exact global maximization.

\subsection{Proof of Exact Selector Sufficiency}
\label{app:proof_exact}

\begin{proof}[Proof of Lemma~\ref{lem:route_exact}]
Fix $z$ and graphs $\mathcal G_1,\mathcal G_2$ satisfying $R_{t,z}(\mathcal G_1)=R_{t,z}(\mathcal G_2)$. Their labeled domains, archived data, prior objects, fitting configurations, and coupled fitting random inputs coincide. By the common builder in Eq.~\eqref{eq:route_factorization}, they therefore produce identical $b,m,s$, hence the same CEI surface. Applying the same measurable numerical selector with common ordering and selection/tie-breaking random inputs yields
\begin{equation}
A_{\mathcal G_1}(z,\xi)=A_{\mathcal G_2}(z,\xi)
\quad\text{for almost every }\xi.
\label{eq:exact_action}
\end{equation}
Their next-intervention laws at this fixed state consequently agree. No existence or uniqueness of a global maximizer is needed. If route equality holds for every admissible $z$, this equality holds at every such state; equality at one state alone makes no claim about other states.
\end{proof}

A known-graph CBO route can serve as the reference under the same state and builder. In fully observed settings this can be the route of $\mathcal G^\star$. With latent variables, the reference must undergo the same observable-variable processing and be a valid input to the same builder; a full latent DAG is not automatically interchangeable with the maintained observed-variable proxy. The result preserves a reference selector, not identification of the true interventional law. Route equality is sufficient, not necessary, since distinct acquisition surfaces can select the same intervention.

\subsection{Proof of Approximate CEI Stability}
\label{app:proof_stability}

Let $\mathcal E$ be the common finite nonempty labeled family and
$\mathcal A=\bigcup_{S\in\mathcal E}\{S\}\times\mathcal D(S)$ the common action space. Fix the two realized rebuilt predictors. For minimization, write $g(a)=b-m(a)$ and $\widetilde g(a)=\widetilde b-\widetilde m(a)$, with real-valued means and incumbents and finite standard deviations $s(a),\widetilde s(a)\ge0$. Assume common finite costs $c(a)\ge c_{\min}>0$ and the finite uniform bounds in Proposition~\ref{prop:cei_stability}. Define
\begin{equation}
I(g,s)=\mathbb E[(g+sU)_+],\qquad
\alpha(a)=\frac{I(g(a),s(a))}{c(a)},\qquad U\sim\mathcal N(0,1),
\label{eq:gaussian_ei_theory}
\end{equation}
and define $\widetilde\alpha$ analogously. Thus $\alpha$ here denotes the complete cost-normalized acquisition in Eq.~\eqref{eq:routed_action}. For maximization the same result holds with $g=m-b$.

\begin{proof}[Proof of Proposition~\ref{prop:cei_stability}]
For $s>0$, the Gaussian integral gives $I(g,s)=g\Phi(g/s)+s\phi(g/s)$, where $\Phi,\phi$ are the standard normal CDF and density. Differentiating this closed form yields $\partial_g I=\Phi(g/s)$ and $\partial_s I=\phi(g/s)$. Their absolute values are bounded by $1$ and $1/\sqrt{2\pi}$, respectively. Integrating along the two coordinate directions gives
\begin{equation}
|I(g,s)-I(\widetilde g,\widetilde s)|
\le |g-\widetilde g|+
\frac{|s-\widetilde s|}{\sqrt{2\pi}}.
\label{eq:ei_lipschitz}
\end{equation}
This also holds when either standard deviation is zero. Indeed, $I(g,0)=g_+$ and the positive-part map is 1-Lipschitz, so
$|I(g,s)-I(g,0)|\le s\,\mathbb E|U|\to0$.
Continuity extends the inequality to the boundary without differentiating there. Dividing by the common $c(a)\ge c_{\min}$ and taking the supremum over $\mathcal A$ proves Eq.~\eqref{eq:cei_stability}.

Let the perturbed selector return $\widetilde a\in\mathcal A$ with
$\widetilde\alpha(\widetilde a)\ge\sup_{a\in\mathcal A}\widetilde\alpha(a)-\zeta$ for finite $\zeta\ge0$. Since its value is finite, this condition entails a finite perturbed supremum; the uniform bound then also makes the reference supremum finite. For every $a\in\mathcal A$,
\begin{equation}
\alpha(a)\le\widetilde\alpha(a)+\eta
\le\widetilde\alpha(\widetilde a)+\zeta+\eta
\le\alpha(\widetilde a)+2\eta+\zeta.
\label{eq:cei_argmax_chain}
\end{equation}
Taking the supremum over $a$ proves the reference-acquisition loss bound, even if the reference supremum is not attained.
\end{proof}

\begin{corollary}[Preservation of the selected ES]
\label{cor:es_margin}
Under Proposition~\ref{prop:cei_stability}, let $M_S=\sup_{\mathbf x\in\mathcal D(S)}\alpha(S,\mathbf x)$ be finite and have a unique best ES $S^\star$. For $|\mathcal E|>1$ define
\begin{equation}
\Gamma=M_{S^\star}-\max_{S\ne S^\star}M_S,
\label{eq:es_margin}
\end{equation}
and set $\Gamma=+\infty$ when $|\mathcal E|=1$. If $\Gamma>2\eta+\zeta$, every $\zeta$-optimal perturbed selection uses $S^\star$.
\end{corollary}
\begin{proof}
The singleton-family case is immediate. Otherwise a returned $\widetilde a=(S,\mathbf x)$ with $S\ne S^\star$ would have reference-acquisition loss at least $M_{S^\star}-M_S\ge\Gamma>2\eta+\zeta$, contradicting Proposition~\ref{prop:cei_stability}. This preserves the ES identity, not necessarily the continuous intervention value within it.
\end{proof}

\subsection{Why Both Routing Components Matter}
\label{app:route_components}

The common-family condition cannot be replaced by agreement only on shared actions. Consider $\mathcal E=\{\{X_1\},\{X_2\},\{X_1,X_2\}\}$ and $\widetilde{\mathcal E}=\{\{X_1\},\{X_2\}\}$, each with singleton domains and unit costs. For any $q,L>0$, take deterministic Gaussian predictions ($s=0$), incumbent $b=0$, and means $-q$ on the two singleton actions and $-(q+L)$ on the joint action. The shared predictions agree exactly, but removing the joint ES incurs reference-CEI loss $L$. This is an optimizer-interface counterexample, not an assertion that every such predictive assignment is induced by a particular fitted SCM.

Conversely, equal ES families alone do not preserve selection. On two actions with unit cost, $b=0$, and $s=0$, exchanging the means $-1$ and $-2$ reverses the maximizing action without changing the family. Uncertainty matters as well: when $g=0$, EI is $s/\sqrt{2\pi}$, so matching prior means alone cannot bound CEI differences. Together these examples explain why routing analysis tracks both the available sets and their predictive scoring.

\paragraph{Relation to structural and prior-work criteria.}
Graph equality is stronger than route equality at matched inputs. Structural discrepancies outside the inputs used by the routing builder can be invisible to CEI, while a single edit removing an available joint intervention can affect selection. Thus structural distance alone does not supply the predictive-error bounds required above. Section~\ref{sec:related} discusses the corresponding prior-work distinctions. Our analysis concerns the joint graph-dependent RuntimeES and rebuilt predictive interface of committed-graph CBO, with an explicit CEI perturbation and ES-margin bound.

\paragraph{Applying the fixed-update guarantees.}
Applying Proposition~\ref{prop:cei_stability} requires uniform error bounds on the rebuilt predictive quantities, including incumbent changes and uncertainty, and a bound $\zeta$ on numerical maximization error. These are analytic conditions on the predictor and selector. The implemented ancestry and $n\mathrm{SHD}$ checks provide graph-commitment proxies, while the fitted-prior-mean diagnostic measures one component of predictive change. Extending this analysis to evolving optimization runs requires controlling subsequent data, refits, archives, and accepted updates.

\end{document}